%% file: main.tex
\documentclass[]{brave}
\usepackage{amssymb}
\usepackage{brave-math}
\usepackage{pifont}
\usepackage[misc]{ifsym}

\usepackage{xspace}
\usepackage{wrapfig}

\makeatletter
\DeclareRobustCommand{\name}{\@ifstar{\name@nospace}{\name@space}}
\newcommand{\name@space}{WorldExam\xspace}
\newcommand{\name@nospace}{WorldExam}
\makeatother
\newcommand{\corrmark}{\text{\normalfont\Letter}}

\title{WorldExam: Benchmarking World Models from Apparent Appearance to Inherent Reactivity}

\author[1, *, \dagger]{Yuxue Yang}
\author[1, *]{Shuyao Shang}
\author[1]{Jiahe Wang}
\author[1]{Zitong Zhou}
\author[1]{Liang Tan}
\authorbreak
\author[1]{Junhan Zeng}
\author[1]{Ruizhi Li}
\author[1]{Junyan Li}
\author[4]{Yu Liu}
\author[5]{Xiao Yang}
\author[5]{Yong Li}
\authorbreak
\author[5]{Jun Zhu}
\author[2,3]{Hongsheng Li}
\author[1]{Tieniu Tan}
\author[1, \dagger,\ \corrmark]{Lue Fan}
\author[1,\ \corrmark]{Zhaoxiang Zhang}

\affiliation[1]{CASIA}
\affiliation[2]{SLAI}
\affiliation[3]{CUHK}
\affiliation[4]{AMAP}
\affiliation[5]{THU}

\contribution[*]{Equal Contribution}
\contribution[\dagger]{Project Leaders}
\contribution[\corrmark]{Corresponding Authors}

\abstract{%
Controllable video generation models are increasingly being developed as world models.
Accordingly, evaluating them in this role extends beyond the \emph{apparent appearance} of generated videos to the \emph{inherent reactivity} of the worlds they depict:
the ability to infer from the scene state how the world should react and to generate plausible consequences not explicitly described in the input.
Yet existing benchmarks mainly assess visual quality or explicit instruction fulfillment by checking whether requested actions and interaction outcomes are realized, leaving inherent reactivity underexamined.
We introduce \textbf{\name}, a hierarchical diagnostic benchmark spanning four levels: Visual Quality, Control Adherence, Spatial Consistency, and World Reactivity.
It comprises 1{,}474 cases across eight dedicated tasks and supports unified evaluation of camera-, action-, and language-driven model paradigms.
The World Reactivity level evaluates scene-conditioned reactions and goal-directed behaviors beyond what is explicitly specified in the input.
Evaluation of 20 representative models reveals a clear capability split.
Camera-driven models excel at camera control, but their interfaces do not support dynamic interaction;
action-driven models control subjects more precisely but often leave the world unresponsive;
and language-driven models perform better on interaction but follow complex controls less faithfully.
No model combines broad task coverage with consistently strong performance, showing that high visual quality and explicit instruction fulfillment do not guarantee inherent reactivity.
}

\checkdata[Project Website]{%
  \url{https://WorldExam.github.io}%
}

\begin{document}

\maketitle


\vspace{-5mm}
\begin{figure}[t]
  \centering
  \includegraphics[width=\linewidth]{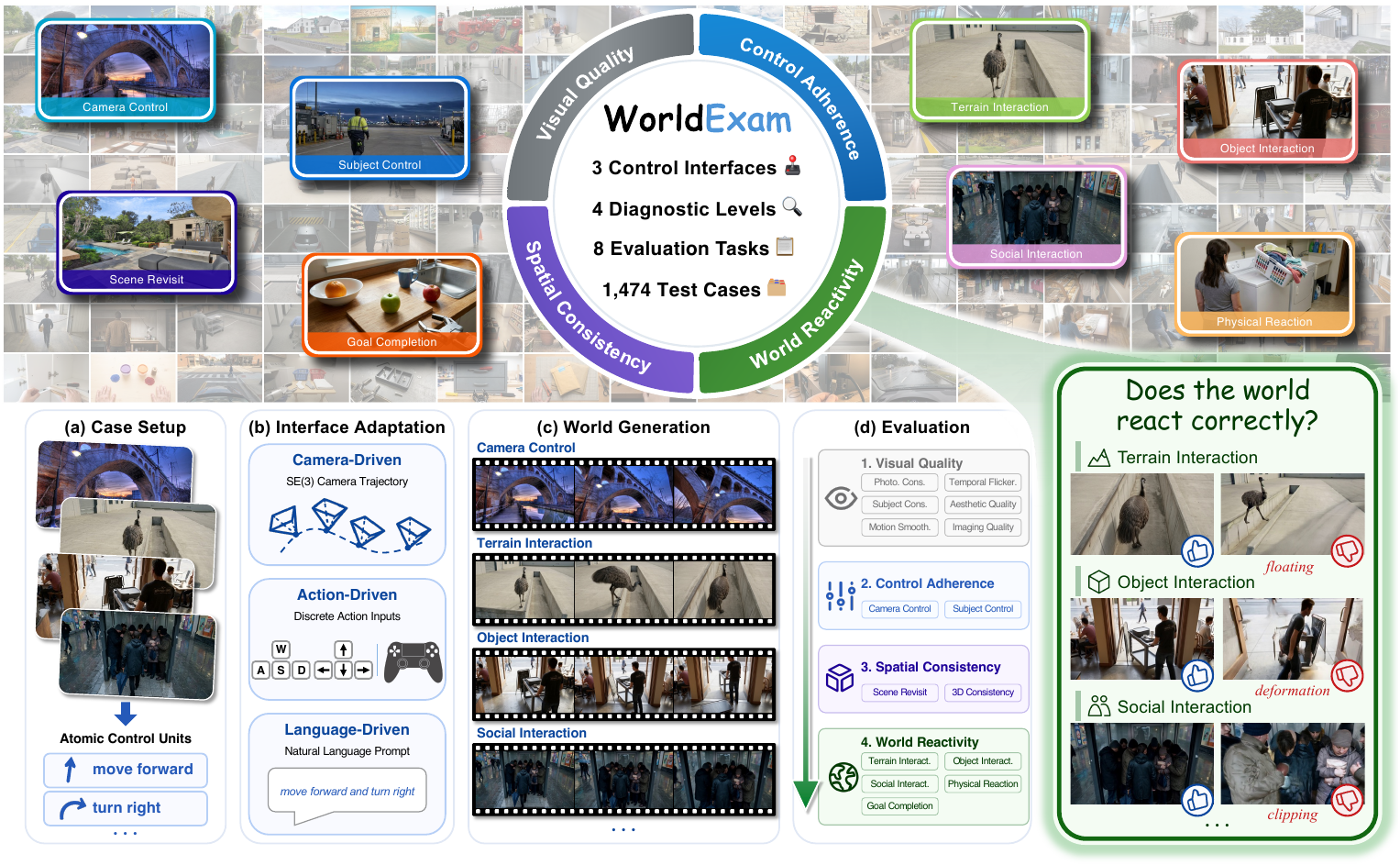}
  \caption{\textbf{Overview of \name.}
  \name is a hierarchical diagnostic benchmark from \emph{apparent appearance} to \emph{inherent reactivity}.
  It evaluates 1,474 test cases across camera-, action-, and language-driven interfaces, four diagnostic levels, and eight tasks under a unified evaluation pipeline.}
  \label{fig:teaser}
\end{figure}

\input{sections/1_introduction}

\input{sections/2_related_work}
\input{sections/3_design}
\input{sections/4_evaluation}
\input{sections/5_experiments}

\input{sections/6_conclusion}

\clearpage
\beginappendix
\input{sections/x_appendix}

\clearpage
\bibliographystyle{plainnat}
\bibliography{main}
\end{document}

%% file: sections/1_introduction.tex
\section{Introduction}
\label{sec:intro}

Controllable video generation models are increasingly being developed as world models rather than standalone clip generators~\citep{brooks2024video,bruce2024genie,yang2023unisim,recammaster,neoverse,worldplay,lingbot,vidu,veo}.
Such models are expected to predict future visual states from an initial observation and control instructions, including camera trajectories, action sequences, and language prompts.
Evaluating them in this role extends beyond the \emph{apparent appearance} of generated videos to the \emph{inherent reactivity} of the worlds they depict~\citep{yang2026mirabench,li2026robotrustbench}.
When a subject moves onto stairs, its motion should adapt to the terrain; when it approaches an obstacle, the world should show contact, avoidance, or blockage; when it enters another agent's personal space, that agent should respond plausibly.
These effects are scene-conditioned consequences rather than direct depictions of the input.
Together, they reveal a model's \emph{inherent reactivity}: its ability to infer from the scene state how the world should react and to generate such consequences plausibly.

Recent benchmarks have advanced world-model evaluation beyond perceptual quality to structured layout control~\citep{duan2025worldscore}, unified action interfaces~\citep{ye2026mind,xu2026worldmark,fang2026iworld,ying2026wbench,xu2026worldroambench}, prompt-specified interaction effects~\citep{wu2026omniworldbench,zhao2026worldolympiad}, and embodied-AI and autonomous-driving applications~\citep{shang2026worldarena,liang2025worldlens}.
As summarized in \cref{tab:benchmark-comparison}, they span camera-, action-, and language-driven model paradigms and increasingly cover camera and subject control, scene revisiting, and interaction outcomes.
Complementary benchmarks probe implicit rules, future-state reasoning, and law-specific physical consistency in specialized settings~\citep{liu2026risevideo,wu2026worldreasonbench,upadhyay2026worldbench,lin2026phyground}.
Yet most benchmarks still assess explicit instruction fulfillment: a desired layout, camera trajectory, action sequence, or interaction consequence is specified in advance, and the model is evaluated on whether the specified outcome is realized.
This evaluation is necessary, but it leaves underexamined a model's ability to infer additional consequences implied by the initial state but not described in the instruction.

We introduce \textbf{\name}, a hierarchical diagnostic benchmark designed around this distinction, as summarized in \cref{fig:teaser}.
\name represents each controllable behavior as a composition of atomic control units and adapts these units to each model's native interface: $\mathrm{SE}(3)$ camera trajectories for camera-driven models, discrete action sequences for action-driven models, and natural-language prompts for language-driven models.
For World Reactivity cases, the model-facing instruction specifies only the explicit control or goal, leaving the expected scene-conditioned reactions unstated.
This design distinguishes direct fulfillment of a requested outcome from behavior beyond what the input explicitly specifies.

\name organizes evaluation into the four diagnostic levels: \emph{Visual Quality}, \emph{Control Adherence}, \emph{Spatial Consistency}, and \emph{World Reactivity}.
We instantiate this hierarchy with eight evaluation tasks: Camera Control, Subject Control, Scene Revisit, Terrain Interaction, Object Interaction, Social Interaction, Physical Reaction, and Goal Completion.
\emph{Visual Quality} is measured with task-agnostic metrics; \emph{Control Adherence} is evaluated through Camera Control and Subject Control; and \emph{Spatial Consistency} is evaluated through Scene Revisit.
The \emph{World Reactivity} level covers scene-conditioned reactions and goal-directed behaviors.
Within this level, four reaction-oriented tasks use control units as triggers while leaving the induced scene-conditioned reactions unstated.
Goal Completion extends the same principle to goal-directed behavior:
it specifies a high-level goal while leaving the detailed execution steps unstated.
For example, a goal to arrange three bolts by height specifies the target layout, but not which object to move first or how to realize the motion frame by frame.

For model-interface compatibility, \name uses two tracks rather than one global ranking.
The static-scene track controls only the camera and is available to all three paradigms.
The dynamic-interaction track requires observable subject--environment interaction and is therefore evaluated only on compatible action- and language-driven models.
Separating the tracks avoids treating unsupported capabilities as failures or averaging scores obtained under different scene assumptions and task sets.

Our evaluation of 20 representative models reveals clear trade-offs across the four levels and three paradigms.
Camera-driven models excel at camera control, but their interfaces do not support dynamic interaction.
Action-driven models control subjects more precisely but often leave the world unresponsive.
Language-driven models perform better on interaction tasks but follow complex controls less faithfully.
These capability splits are obscured by aggregate scores, motivating separate reporting at both level and task granularity.

Our contributions are summarized as follows.
\begin{itemize}\setlength{\itemsep}{1pt}
  \item We extend world model evaluation beyond apparent appearance to \emph{inherent reactivity}: inferring from the scene state how the world should react and generating plausible consequences absent from the input.
  \item We propose \name, a benchmark of 1{,}474 cases across eight tasks that supports unified evaluation of camera-, action-, and language-driven model paradigms.
  \item We evaluate 20 representative models, revealing paradigm-dependent capability splits. No model combines broad task coverage with consistently strong performance, showing that high visual quality and explicit instruction fulfillment do not guarantee inherent reactivity.
  \item We will publicly release the benchmark data and evaluation toolkit to facilitate systematic evaluation and foster continued progress in the video world model community.
\end{itemize}

\input{tables/benchmark_comparison}

%% file: tables/benchmark_comparison.tex
\begingroup
\newcommand{\benchyes}{\textcolor{green!55!black}{\ding{51}}}
\newcommand{\benchno}{\textcolor{red!70!black}{\ding{55}}}
\newcommand{\pcam}{\textcolor{blue!70!black}{\textsf{C}}}
\newcommand{\pact}{\textcolor{orange!85!black}{\textsf{A}}}
\newcommand{\plang}{\textcolor{violet!80!black}{\textsf{L}}}
\begin{table*}[t]
\centering
\caption{\textbf{Comparison with representative world-model benchmarks.}
The table compares supported model paradigms, viewpoints, task coverage, case counts, and evaluated models.
\pcam{}, \pact{}, and \plang{} denote camera-, action-, and language-driven model paradigms, respectively.
\textsuperscript{\ensuremath{\dagger}} indicates that the instruction specifies the expected interaction consequence; the corresponding \name tasks leave the evaluated reaction unstated.
}
\label{tab:benchmark-comparison}
\scriptsize
\setlength{\tabcolsep}{2.7pt}
\renewcommand{\arraystretch}{1.18}
\resizebox{\textwidth}{!}{%
\begin{tabular}{@{}l*{11}{c}rr@{}}
\toprule
\multirow{2}{*}[-6pt]{\textbf{Benchmark}} &
\multirow{2}{*}[-6pt]{\shortstack[c]{\textbf{Model}\\\textbf{Paradigm}}} &
\multicolumn{2}{c}{\textbf{Viewpoint}} &
\multicolumn{8}{c}{\textbf{\name Evaluation Tasks}} &
\multirow{2}{*}[-6pt]{\textbf{\#Cases}} &
\multirow{2}{*}[-6pt]{\textbf{\#Models}} \\
\cmidrule(lr){3-4}\cmidrule(lr){5-12}
&
&
\shortstack{\textbf{First}\\\textbf{Person}} &
\shortstack{\textbf{Third}\\\textbf{Person}} &
\shortstack{\textbf{Camera}\\\textbf{Control}} &
\shortstack{\textbf{Subject}\\\textbf{Control}} &
\shortstack{\textbf{Scene}\\\textbf{Revisit}} &
\shortstack{\textbf{Terrain}\\\textbf{Inter.}} &
\shortstack{\textbf{Object}\\\textbf{Inter.}} &
\shortstack{\textbf{Social}\\\textbf{Inter.}} &
\shortstack{\textbf{Physical}\\\textbf{React.}} &
\shortstack{\textbf{Goal}\\\textbf{Compl.}} &
&
\\
\midrule
WorldScore~\citep{duan2025worldscore}
& \pcam/\plang & \benchyes & \benchno
& \benchyes & \benchno & \benchno & \benchno & \benchno & \benchno & \benchno & \benchno
& 3,000 & 20 \\
MIND~\citep{ye2026mind}
& \pact & \benchyes & \benchyes
& \benchyes & \benchyes & \benchyes & \benchno & \benchno & \benchno & \benchno & \benchno
& 250 & 2 \\
Omni-WorldBench~\citep{wu2026omniworldbench}
& \pcam/\plang & \benchyes & \benchyes
& \benchyes & \benchyes & \benchyes & \benchno & \benchyes\textsuperscript{\ensuremath{\dagger}} & \benchno & \benchyes\textsuperscript{\ensuremath{\dagger}} & \benchno
& 1,068 & 18 \\
WorldMark~\citep{xu2026worldmark}
& \pcam/\pact/\plang & \benchyes & \benchyes
& \benchyes & \benchno & \benchno & \benchno & \benchno & \benchno & \benchno & \benchno
& 500 & 6 \\
iWorld-Bench~\citep{fang2026iworld}
& \pcam/\pact/\plang & \benchyes & \benchno
& \benchyes & \benchno & \benchyes & \benchno & \benchno & \benchno & \benchno & \benchno
& 4,900 & 14 \\
WBench~\citep{ying2026wbench}
& \pcam/\pact/\plang & \benchyes & \benchyes
& \benchyes & \benchyes & \benchyes & \benchno & \benchyes\textsuperscript{\ensuremath{\dagger}} & \benchno & \benchyes\textsuperscript{\ensuremath{\dagger}} & \benchno
& 289 & 20 \\
WorldOlympiad~\citep{zhao2026worldolympiad}
& \pact/\plang & \benchyes & \benchno
& \benchyes & \benchno & \benchno & \benchno & \benchyes\textsuperscript{\ensuremath{\dagger}} & \benchno & \benchyes\textsuperscript{\ensuremath{\dagger}} & \benchno
& 1,000 & 8 \\
WorldRoamBench~\citep{xu2026worldroambench}
& \pact & \benchyes & \benchyes
& \benchyes & \benchyes & \benchyes & \benchyes & \benchyes & \benchno & \benchyes & \benchno
& 600 & 10 \\
\midrule
\rowcolor{gray!15}
\textbf{\name (Ours)}
& \pcam/\pact/\plang & \benchyes & \benchyes
& \benchyes & \benchyes & \benchyes & \benchyes & \benchyes & \benchyes & \benchyes & \benchyes
& \textbf{1,474} & \textbf{20} \\
\bottomrule
\end{tabular}%
}
\end{table*}
\endgroup

%% file: sections/2_related_work.tex
\section{Related Work}
\label{sec:related}

\subsection{Video World Models}
Recent video world models increasingly support controllable video generation for gaming, robotics, embodied AI, and open-world simulation.
Based on their primary control interfaces, they can be broadly grouped into camera-, action-, and language-driven paradigms.
Camera-driven models~\citep{trajectorycrafter,recammaster,voyager,fantasyworld,neoverse,inspatio} condition generation on camera trajectories, represented in two main ways.
Some approaches, such as ReCamMaster~\citep{recammaster} and FantasyWorld~\citep{fantasyworld}, inject camera trajectories through learned camera encoders or embeddings, whereas others~\citep{trajectorycrafter,voyager,neoverse,inspatio} reconstruct 3D priors from the input and reproject them to target viewpoints; representative methods include NeoVerse~\citep{neoverse} and InSpatio-World~\citep{inspatio}.
Action-driven models~\citep{gamecraft,astra,worldplay,yume15,lingbot,infiniteworld,matrixgame3} generate future frames conditioned on discrete action sequences through keyboard-like interfaces.
Among them, WorldPlay~\citep{worldplay} and LingBot-World~\citep{lingbot} focus on real-time interaction and consistent generation under direct action control.
Language-driven models~\citep{kling,veo,hailuo,wan,seedance,vidu,happyhorse} generate videos from text or image-text prompts, demonstrating advances in semantically complex video generation.
Across paradigms, video world models are evolving from short open-loop synthesis toward controllable, persistent, and interactive environment simulation.
Heterogeneous interfaces complicate direct comparison, while controllability, long-term memory, and \emph{inherent reactivity} remain key challenges.

\subsection{Video World Model Benchmarks}

A growing body of benchmarks evaluates complementary aspects of video world modeling.
Some emphasize perceptual and temporal quality~\citep{huang2024vbench,huang2025vbenchpp,zheng2025vbench2,liu2024evalcrafter,liu2023fetv};
others target compositionality, world knowledge, implicit rules, and future-state reasoning~\citep{sun2024t2vcompbench,chen2025t2vworldbench,liu2026risevideo,wu2026worldreasonbench}.
Physics-oriented benchmarks~\citep{bansal2024videophy,meng2024phygenbench,li2025worldmodelbench,upadhyay2026worldbench,lin2026phyground,xue2026acwmphys,wu2026pdibench} diagnose law-specific dynamics, geometric consistency, and generalization under physical interactions;
embodied benchmarks~\citep{qin2024worldsimbench,yue2025ewmbench,li2025worldeval,shang2026worldarena,jiang2026robowmbench,yang2026mirabench,li2026robotrustbench,liu2026kinebench} evaluate action fidelity, physical executability, planning utility, reliability, and trustworthiness;
and autonomous-driving benchmarks~\citep{arai2024actbench,liang2025worldlens,zhou2026drivinggen} emphasize ego-action control, trajectory plausibility, safety, and downstream driving utility.
Beyond these settings, general benchmarks~\citep{duan2025worldscore,ye2026mind,wu2026omniworldbench,xu2026worldmark,fang2026iworld,ying2026wbench,zhao2026worldolympiad,xu2026worldroambench,zhang2025worldinworld} evaluate interactive world models across varied scenes and interfaces.

Among these general benchmarks, WorldScore~\citep{duan2025worldscore} evaluates camera-trajectory-based layout control and geometric consistency, while MIND~\citep{ye2026mind} focuses on action control and closed-loop revisit consistency.
WorldMark~\citep{xu2026worldmark} and iWorld-Bench~\citep{fang2026iworld} improve cross-model comparison through standardized or unified action representations.
Omni-WorldBench~\citep{wu2026omniworldbench} evaluates prompt-specified interaction outcomes, affected and unaffected entities, and intermediate causal state transitions; WBench~\citep{ying2026wbench} extends evaluation to multi-turn navigation, subject actions, event editing, and perspective switching; and WorldOlympiad~\citep{zhao2026worldolympiad} probes long-horizon interaction and physics.
WorldRoamBench~\citep{xu2026worldroambench} further couples long-horizon action-conditioned generation with diagnostics of controllability, visual drift, mechanics, optics, 3D consistency, and memory.
Collectively, these benchmarks substantially broaden interactive evaluation, but most still center on explicit instruction fulfillment by checking whether specified controls or interaction outcomes are realized.
In contrast, \name adapts atomic control units to each model's native interface and evaluates \emph{inherent reactivity} through scene-conditioned reactions and goal-directed behaviors beyond what the input explicitly specifies.

%% file: sections/3_design.tex
\section{\name}
\label{sec:design}

\name supports unified evaluation of video world models with different control interfaces.
We formulate a video world model as a function $f : \mathcal{I} \times \mathcal{C} \rightarrow \mathcal{V}$, where $\mathcal{I}$ is the initial image, $\mathcal{C}$ is the model-facing input instruction and $\mathcal{V}$ is the generated video.
We consider three common paradigms: \emph{camera-driven} models take camera trajectories in $\mathrm{SE}(3)$, \emph{action-driven} models take discrete action sequences over \{W (move forward), S (move backward), A (move left), D (move right), $\uparrow$ (tilt up), $\downarrow$ (tilt down), $\leftarrow$ (pan left), $\rightarrow$ (pan right), $\varnothing$ (stop)\}, and \emph{language-driven} models take natural-language prompts.

To compare these paradigms, \name uses interface adaptation to map a shared case to each model's native interface.
\name represents controllable behavior as an ordered composition of atomic control units, such as moving forward (W) and then panning right ($\rightarrow$), and adapts this control intent into an $\mathrm{SE}(3)$ camera trajectory, a discrete action sequence, or a natural-language prompt.
Under this setup, \name first defines a four-level diagnostic hierarchy (\cref{sec:design:hierarchy}) and then instantiates it through eight evaluation tasks (\cref{sec:design:task}).
We next describe the test case curation pipeline (\cref{sec:design:curation}) and the benchmark statistics (\cref{sec:design:stats}).
Metric definitions and scoring protocols are described in \cref{sec:eval}.

\subsection{Toward World Reactivity: A Four-Level Diagnostic Hierarchy}
\label{sec:design:hierarchy}
\name organizes world-model evaluation into four diagnostic levels of increasing scope: \emph{Visual Quality}, \emph{Control Adherence}, \emph{Spatial Consistency}, and \emph{World Reactivity}.
\emph{Visual Quality} measures the video's \emph{apparent appearance}, including perceptual plausibility, temporal stability, and aesthetic quality.
\emph{Control Adherence} measures whether the controlled camera or subject follows the input control.
\emph{Spatial Consistency} measures whether the model preserves a coherent world when the camera revisits a previously observed viewpoint.

By contrast, the \emph{World Reactivity} level evaluates scene-conditioned reactions and goal-directed behaviors beyond what is explicitly specified in the input.
In reaction-oriented cases, a control specifies the initiating behavior but leaves its scene-conditioned consequences unstated, which the model must infer from the scene.
For example, a move-forward control specifies the subject's direction but not how its motion should adapt to the terrain.
If there is an obstacle or a nearby agent in its motion path, the subject may stop or avoid it, another agent may yield, or an object may move on contact.
In goal-directed cases, a high-level goal specifies the desired target, and the model needs to infer from the initial scene how to realize it frame by frame.

Although the four levels form a diagnostic progression, strong \emph{Visual Quality}, \emph{Control Adherence}, and \emph{Spatial Consistency} do not guarantee successful scene-conditioned reactions or goal execution.
Conversely, success on \emph{World Reactivity} does not compensate for visual artifacts, control errors, or spatial drift.
\name therefore reports the four levels separately to localize failures in generation quality, explicit control, spatial persistence, and behavior that must be inferred from the scene.

\begin{figure*}[t]
  \centering
  \includegraphics[width=\linewidth]{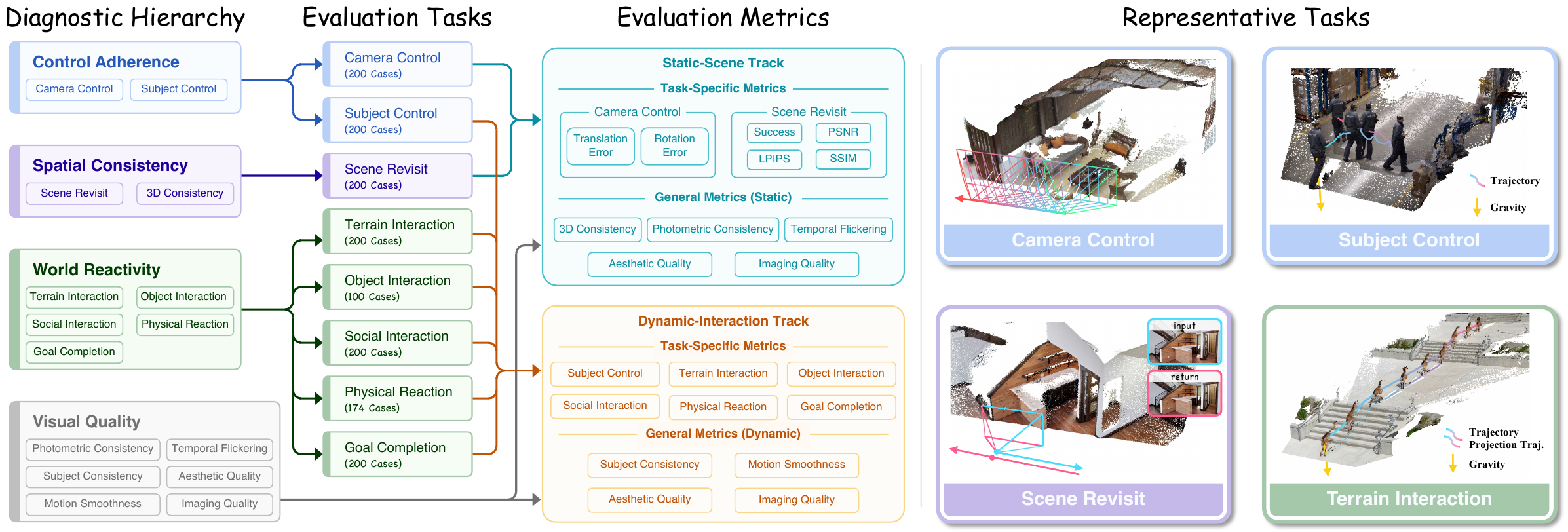}
  \caption{\textbf{\name taxonomy, tracks, and metrics.}
  Four diagnostic levels map to eight evaluation tasks, which are assigned to static-scene or dynamic-interaction tracks according to scene assumptions and model applicability.
  Each track reports task-specific and general metrics.
  Representative examples illustrate the geometry-based evaluations.}
  \label{fig:task-taxonomy}
\end{figure*}

\subsection{From Diagnostic Levels to Evaluation Tasks}
\label{sec:design:task}

\Cref{fig:task-taxonomy} shows how the four diagnostic levels are instantiated.
\emph{Visual Quality} uses task-agnostic metrics across all videos, while the other three levels are instantiated by eight tasks.
\emph{Control Adherence} includes \textbf{Camera Control} and \textbf{Subject Control}, while \emph{Spatial Consistency} uses \textbf{Scene Revisit}.
\emph{World Reactivity} includes four reaction-oriented tasks, \textbf{Terrain Interaction}, \textbf{Object Interaction}, \textbf{Social Interaction}, and \textbf{Physical Reaction}, plus \textbf{Goal Completion} for goal-directed execution.

The tasks are reported through two tracks rather than a single global score to avoid penalizing models for tasks their interfaces do not support.
The \emph{static-scene track} contains Camera Control and Scene Revisit and is available to all three model paradigms because each interface can express camera motion.
The \emph{dynamic-interaction track} contains Subject Control and the five \emph{World Reactivity} tasks and applies only to compatible action- and language-driven models; Goal Completion is language-only.

\paragraph{Control Adherence.}
\textbf{Camera Control} tests whether the generated camera motion follows the prescribed controls.
Each case composes one to three atomic camera controls from \{W, S, A, D, $\uparrow$, $\downarrow$, $\leftarrow$, $\rightarrow$\}, assigns each an execution-time fraction, and executes them in order over the assigned intervals.
\textbf{Subject Control} applies the same construction to a designated third-person subject using \{W, S, A, D\}.

\paragraph{Spatial Consistency.}
\textbf{Scene Revisit} tests the model's spatial memory of the initial observation.
Each case uses a round-trip camera trajectory formed by an outgoing control and its inverse, such as ``move left'' followed by ``move right'', or ``tilt up'' followed by ``tilt down''.
After moving away, the camera should return to the initial viewpoint while the returned view preserves the scene's geometry, appearance, and content.

\paragraph{World Reactivity.}
The four reaction-oriented tasks pair an initial scene with a single atomic subject control.
\textbf{Terrain Interaction} places stairs, slopes, bridges, trenches, or other structured terrain along the controlled subject's path.
The input specifies only the horizontal motion direction, while the model must infer how the subject should adapt its height and maintain contact with the terrain.
\textbf{Object Interaction} places a movable, flexible, or rigid target along the subject's path so that the subject, one of its body parts, or a carried tool is expected to make contact with it.
It evaluates whether the target produces an immediate type-appropriate response, such as motion when loose, deformation when flexible, or blockage when rigid, without interpenetration.
\textbf{Social Interaction} places other agents along the subject's path or within its social distance, creating an imminent local conflict.
It evaluates whether the affected agents respond plausibly through avoidance, yielding, stopping, or changing path.
\textbf{Physical Reaction} tests whether a dynamic process unfolds over time according to physical regularities, including gravity, friction, momentum transfer, constrained motion, fluid response, and pendulum-like swinging.
Each case uses one control from \{W, S, A, D, $\varnothing$ (stop)\}.
A motion control may trigger the process, whereas $\varnothing$ is used when the initial scene is expected to evolve autonomously without subject motion.
Although Object Interaction and Physical Reaction may both involve contact, the former targets the immediate type-conditioned response of a designated object, whereas the latter targets the temporal evolution of a physical process.
\textbf{Goal Completion} is language-only and provides a high-level goal together with an initial scene containing relevant entities, distractors, preconditions, and constraints.
Unlike the four reaction-oriented tasks, it uses no atomic control sequence or execution-time fractions.
The input may state necessary subgoals or ordering constraints.
The model should ground the goal in the initial scene, select the correct entities, ignore distractors, and produce coherent execution steps toward the desired target frame by frame.
\Cref{sec:appendix:task-examples} provides examples of all eight tasks and representative checklists.

\subsection{Test Case Curation Pipeline}
\label{sec:design:curation}

\begin{figure*}[t]
  \centering
  \includegraphics[width=\linewidth]{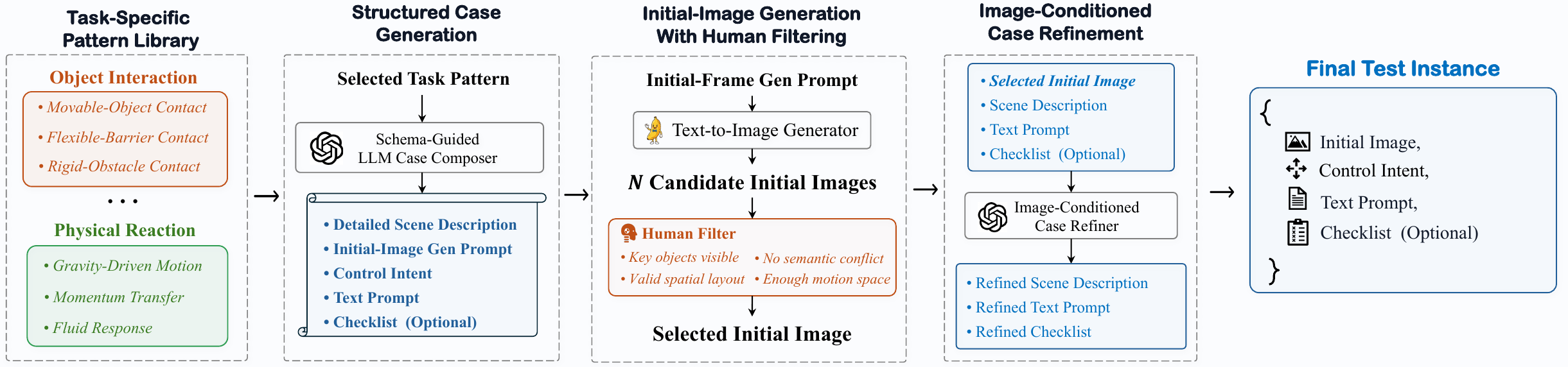}
  \caption{\textbf{Test case curation pipeline.}
  For the six dynamic-interaction tasks, a task pattern is expanded into a structured draft, candidate initial images are generated and human-filtered, and the scene description, text prompt, and optional checklist are refined against the selected image before the case is finalized.}
  \label{fig:test-case-curation-pipeline}
\end{figure*}

\name constructs cases differently for the two tracks.
For static-scene Camera Control and Scene Revisit, we pair suitable first-person scenes from existing datasets~\citep{Flickr2K, dl3dv, ye2026mind} with compositions of atomic control units.
For dynamic-interaction Subject Control and the five World Reactivity tasks, the pipeline in \cref{fig:test-case-curation-pipeline} constructs initial scenes supporting the intended behavior and evaluation.

For each dynamic-interaction task, a task-specific pattern library defines the intended semantic coverage over subject motion, structured terrain, object contact, social conflict, physical processes, or goal-directed situations.
After a pattern is sampled, a schema-guided LLM case composer expands it into a structured draft containing a detailed scene description, an initial-image generation prompt, a control intent or high-level goal, and a draft text prompt for language-driven models.
For Object Interaction, Social Interaction, Physical Reaction, and Goal Completion, the draft also includes a case-specific checklist of observable evaluation criteria.
In each World Reactivity case, the model-facing input specifies only the explicit control or high-level goal; the scene-conditioned reaction or detailed execution process remains unstated.

The initial-image generation prompt is used only to synthesize $N$ candidate initial images.
Human filtering retains candidates in which the relevant entities are visible, the spatial layout supports the intended behavior or event, the image is consistent with the draft, and sufficient motion space remains for the continuation.
Candidates that already depict the evaluated event or desired target, hide relevant entities, or make the intended behavior physically infeasible are discarded.
The selected initial image $I^0$ therefore provides a concrete pre-event state from which the intended behavior, reaction, or goal-directed execution can unfold.

An image-conditioned case refiner then revises the draft to match $I^0$ while preserving the sampled pattern and intended control or goal.
It updates entity references, spatial relations, the scene description, the text prompt, and the optional checklist so that all referenced entities and preconditions are grounded in the selected image.
For tasks evaluated using checklists, the final checklist $L=\{\ell_k\}_{k=1}^{K}$ is fixed at this stage.
The finalized case consists of $I^0$, the control intent or high-level goal, the grounded text prompt, the optional checklist.

\begin{figure}[t]
  \centering
  \begin{subfigure}{\linewidth}
    \centering
    \includegraphics[width=\linewidth]{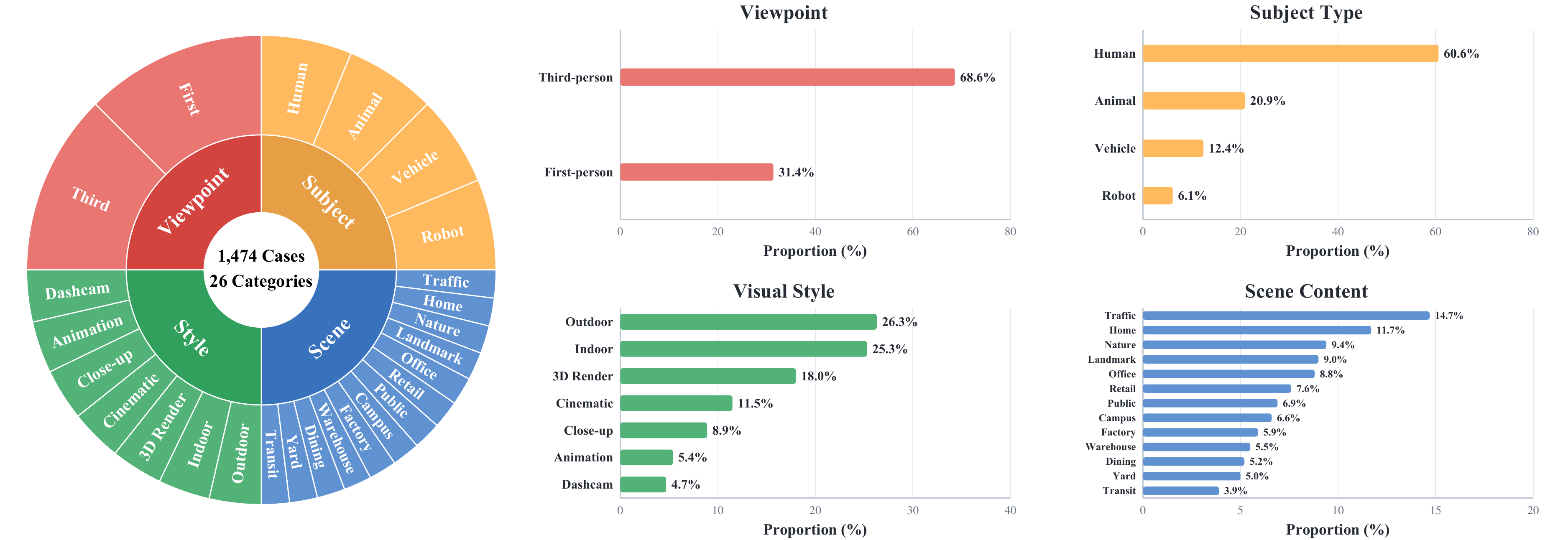}
    \caption{Dataset composition.}
    \label{fig:dataset-composition-overview}
  \end{subfigure}
  \vspace{1mm}
  \begin{subfigure}{\linewidth}
    \centering
    \includegraphics[width=\linewidth]{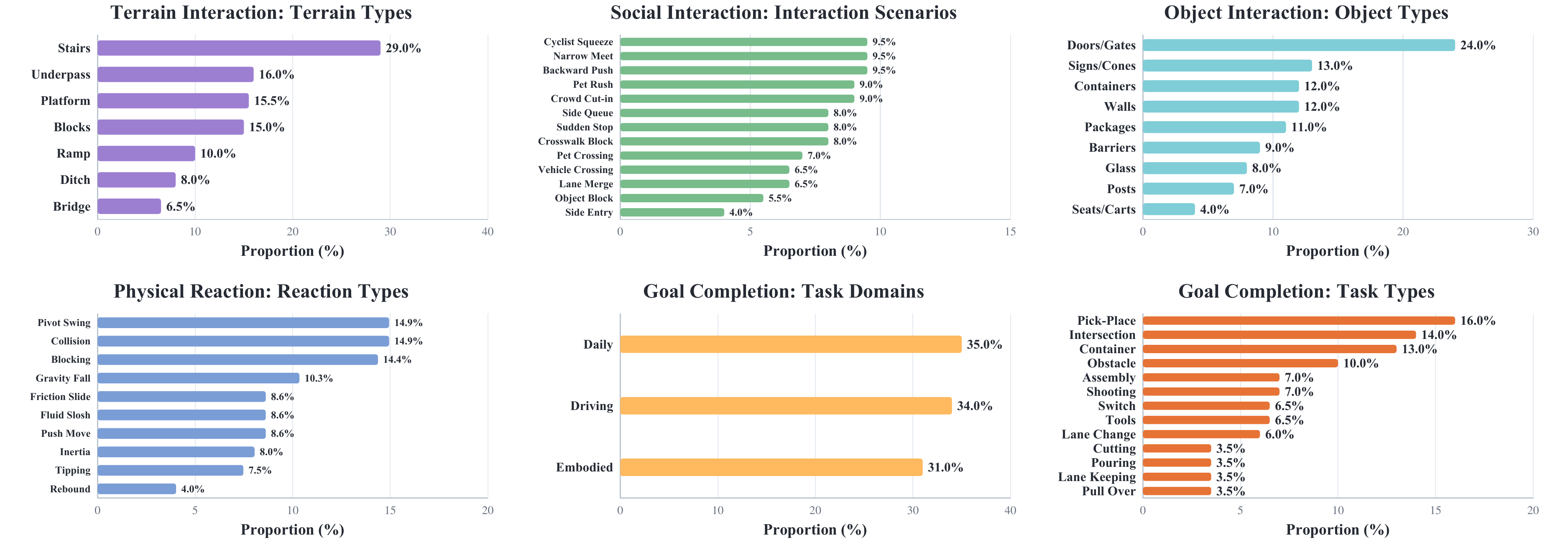}
    \caption{World Reactivity composition.}
    \label{fig:task-breakdown-bars}
  \end{subfigure}
  \caption{\textbf{Benchmark statistics.}
  The top panel summarizes distributions by viewpoint, subject type, visual style, and scene content.
  The bottom panel shows terrain types, social-interaction scenarios, object types, physical-reaction types, and Goal Completion domains and task types.}
  \label{fig:benchmark-statistics}
\end{figure}

\subsection{Benchmark Statistics}
\label{sec:design:stats}
As shown in \cref{fig:task-taxonomy}, \name contains 1{,}474 cases across eight evaluation tasks.
\Cref{fig:benchmark-statistics} summarizes both the overall dataset composition and the task-specific composition of the five \emph{World Reactivity} tasks.

At the dataset level, the cases span first- and third-person viewpoints with first-person viewpoints accounting for 31.4\% of the benchmark and providing substantial egocentric coverage.
The subject taxonomy spans humans, animals, vehicles, and robots, while the visual-style taxonomy mixes outdoor and indoor real scenes with 3D renderings, cinematic footage, close-up views, animation, and dashcam videos.
Scene content is also deliberately broad: no single scene type dominates the benchmark, and the largest category, traffic scenes, accounts for only 14.7\% of the cases.

Within the five \emph{World Reactivity} tasks, the cases are further distributed across task-specific semantic subcategories, including terrain types, social-interaction scenarios, object types, physical-reaction types, and Goal Completion domains and task types.
No single subcategory accounts for more than 35\% of its corresponding task.
This coverage reduces dependence on any one visual or semantic template and supports task-specific analysis across diverse scene-conditioned reactions and goal-directed situations.
Representative cases across these dimensions are shown in \cref{sec:appendix:gallery}.

%% file: sections/4_evaluation.tex
\section{Evaluation Protocol and Metrics}
\label{sec:eval}
For Camera Control, Scene Revisit, Subject Control, and Terrain Interaction, we lift each generated video into 3D with a geometry reconstruction model and evaluate it in the reconstructed space.
Camera Control and Scene Revisit use the recovered camera trajectories, whereas Subject Control and Terrain Interaction use the recovered 3D subject trajectories and terrain geometry.
For the remaining four tasks, we use GPT-5.5 as the vision-language model (VLM) judge to score generated videos against predefined case-specific checklists.
Each track reports task-specific metrics together with task-agnostic general metrics for visual quality.

\subsection{Static-Scene Track}
\label{sec:eval:geometry}

Given generated frames $V=\{I_t\}_{t=1}^{T}$, we use VGGT-$\Omega$~\citep{wang2026vggt} to estimate camera poses, intrinsics, and depths.

\paragraph{\textbf{Camera Control}.}
Each case specifies an ordered sequence $\{(a_i,\rho_i)\}_{i=1}^{N} (1\leq N\leq 3)$.
Here, $a_i\in$ \{W, S, A, D, $\uparrow$, $\downarrow$, $\leftarrow$, $\rightarrow$\} is an atomic control unit, and $\rho_i$ is its execution-time fraction, with $\sum_i\rho_i=1$.
For camera- and action-driven interfaces, we allocate $n_i = \left[\rho_i T\right]$ frames to the $i$-th control using nearest-integer rounding, and adjust the allocation to ensure $\sum_i n_i=T$.
From each generated video, we recover a frame-wise camera trajectory $\{(\mathbf{R}_t,\mathbf{t}_t)\}_{t=1}^{T}$.
Because the three interfaces specify camera motion differently, we construct the model-facing input and evaluation reference separately for each interface.

For camera-driven models, controls are composed sequentially starting from the initial camera pose, with each subsequent control applied relative to the endpoint pose of the previous control.
These endpoints serve as keyframes, which we interpolate over the allocated frame intervals to obtain a frame-wise input trajectory in $\mathrm{SE}(3)$.
This input trajectory also serves directly as the frame-wise reference trajectory $\{(\mathbf{R}_t^*,\mathbf{t}_t^*)\}_{t=1}^{T}$.
Before comparison, we express both the recovered and reference trajectories relative to their respective first-frame poses.
We then compute the translation and rotation errors $(e_t, e_r)$ between the two trajectories as
\begin{equation}
e_t = \min_{s\geq 0}\frac{1}{T}\sum_{t=1}^{T}
\left\|s\mathbf{t}_t-\mathbf{t}_t^*\right\|_2,
\qquad
e_r = \frac{180}{\pi}\frac{1}{T}\sum_{t=1}^{T}
\arccos\!\left(
\frac{\operatorname{tr}\!\left(\mathbf{R}_t(\mathbf{R}_t^*)^\top\right)-1}{2}
\right).
\end{equation}
The nonnegative scale $s$ resolves the translation-scale ambiguity of monocular camera reconstruction, making $e_t$ scale-invariant, while $180/\pi$ converts $e_r$ from radians to degrees.
In implementation, the argument of $\arccos$ is clipped to $[-1,1]$ for numerical stability.

Action-driven models map each control to the model's native discrete action and assign the corresponding $n_i$ frames to the $i$-th action.
Language-driven models instead verbalize each control, join the resulting motion phrases in order with ``then,'' and prepend the instruction to the scene description; for example, ``W'' followed by ``$\rightarrow$'' becomes ``The camera moves forward, then pans right. [Scene description].''
This prompt preserves the control order but does not specify the duration of each control.

Unlike camera-driven interfaces, action- and language-driven interfaces do not specify an exact camera trajectory in $\mathrm{SE}(3)$, so we evaluate their recovered trajectories against control-level references segment by segment.
For action-driven models, the frame ranges assigned to the discrete actions directly define the segment boundaries.
For language-driven models, we instead partition the recovered trajectory into $N$ segments by applying dynamic-programming-based change-point detection~\citep{ruptures} to frame-to-frame changes in translation and rotation.
The resulting segments are matched in temporal order to the $N$ atomic controls.

Within each segment, we express the recovered camera poses relative to the first frame, so that the segment starts from the identity pose.
The assigned control determines whether the reference motion is a translation or a rotation.
For a translation control, we linearly interpolate the reference translation from $\mathbf{0}$ to a unit vector $\mathbf{u}_i$ in the prescribed direction, while keeping $\mathbf{R}^*=\mathbf{I}$ throughout.
The unit displacement is sufficient because $e_t$ is invariant to translation scale.
For a rotation control, we set $\mathbf{t}^*=\mathbf{0}$ and construct $\mathbf{R}^*$ using the prescribed axis and direction; its translation error is computed as the mean per-frame $\|\mathbf{t}_t\|_2$ without scale alignment.
Because the interface does not specify a rotation angle, the reference angle is linearly interpolated from zero to the total angle recovered within the segment.
The segment-level references therefore evaluate the prescribed direction and motion progression without imposing a fixed magnitude.

For each segment $i$, we compute $(e_{t,i},e_{r,i})$ using the error definitions above.
A translation segment is assigned the maximum translation error $e_{t,i}=0.5$ if its displacement is below 5\% of the largest segment displacement in the same video or if its net motion is not aligned with the prescribed direction.
A rotation segment is assigned the maximum rotation error $e_{r,i}=15^{\circ}$ if its total rotation angle is below $5^{\circ}$ or is opposite to the prescribed direction.
Finally, we obtain the video-level errors $(e_t,e_r)$ by averaging the corresponding segment-level errors using the number of frames in each segment as weights.

Across all interfaces, we normalize the resulting errors as $s_t=\max(0,1-e_t/0.5)$ and $s_r=\max(0,1-e_r/15)$, where $e_r$ is measured in degrees, and report their geometric mean,
$S_{\mathrm{cam}}=100\sqrt{s_t s_r}$.
Thus, a high Camera Control score requires the recovered camera motion to follow the prescribed directions and temporal progression.

\paragraph{\textbf{Scene Revisit}.}
Scene Revisit evaluates two requirements after a round-trip camera motion: returning the camera to its initial pose and preserving the initial scene in the returned view.
Each case pairs an outgoing control $a_1$ with its inverse $a_2=a_1^{-1}$.
We set their execution-time fractions to $\rho_1=0.4$ and $\rho_2=0.6$, reserving a longer temporal window for the return motion so that the model has sufficient opportunity to reach the initial viewpoint.
We adapt this pair to the three interfaces as in Camera Control.

Let $P_t=(\mathbf{R}_t,\mathbf{t}_t)$ denote the recovered camera pose.
For camera- and action-driven models, $\mathcal{T}_{\mathrm{return}}$ is the frame range allocated to $a_2$; for language-driven models, whose prompt does not specify control duration, it begins at 40\% of the video.
Because the camera may return before the video ends, we search this entire segment and select the frame whose recovered pose is closest to the initial pose:
\begin{equation}
t_{\mathrm{rev}}
=
\underset{t\in\mathcal{T}_{\mathrm{return}}}{\arg\min}\
d(P_t,P_1).
\end{equation}
For translation round trips, $d(P_t,P_1)=\|\mathbf{t}_t-\mathbf{t}_1\|_2$; for rotation round trips, $d$ is the relative rotation angle between $\mathbf{R}_t$ and $\mathbf{R}_1$.
A translation revisit succeeds when this minimum distance is within 10\% of the maximum displacement reached during the outgoing segment; a rotation revisit succeeds when its minimum angular distance is below $5^{\circ}$.
Averaging this binary result over all cases gives Revisit Success $S_{\mathrm{succ}}\in[0,1]$.

We then compare the input image with the selected revisit frame using PSNR, LPIPS, and SSIM.
Selecting the frame by recovered pose rather than using the final frame makes this appearance comparison insensitive to small differences in return timing.
After averaging over cases, we normalize the three appearance metrics as $s_{\mathrm{P}}=\min(\mathrm{PSNR}/25,1)$, $s_{\mathrm{L}}=1-\mathrm{LPIPS}$, and $s_{\mathrm{S}}=\mathrm{SSIM}$, and combine them with Revisit Success:
\begin{equation}
S_{\mathrm{rev}}
=
100\sqrt{
S_{\mathrm{succ}}\cdot\frac{s_{\mathrm{P}}+s_{\mathrm{L}}+s_{\mathrm{S}}}{3}
}.
\end{equation}
The geometric aggregation gives a high Scene Revisit score only when the camera both returns to the initial viewpoint and recovers a consistent view.

\paragraph{\textbf{General metrics}.}
Across all static-scene videos, we additionally report five task-agnostic general metrics.
3D Consistency adapts the metric of WorldScore~\citep{duan2025worldscore} to VGGT-$\Omega$ outputs.
Using the recovered geometry, we project valid pixels from a source frame to a nearby target and back, then measure the cycle reprojection error.
Photometric Consistency measures the forward-backward optical-flow cycle error between neighboring frames using average endpoint error (AEPE).
Temporal Flickering, Aesthetic Quality, and Imaging Quality are adapted from VBench~\citep{huang2024vbench}.
These metrics summarize whether the static-scene generations are geometrically stable, temporally coherent, and visually plausible.

\subsection{Image-Space Displacement Alignment for Camera-Driven Models}
\label{sec:eval:scale-alignment}
Identical translation values in an input $\mathrm{SE}(3)$ trajectory can induce different image-space displacements across camera-driven models because their pose-conditioning interfaces interpret the translation magnitude differently.
Larger displacements expose more novel-view content and increase the difficulty of generating controlled, spatially consistent videos.
By increasing the input translation multiplier to induce progressively larger image-space displacements, the results in \cref{tab:camera-scale-alignment-ablation,fig:scale-ablation} show corresponding declines in Camera Control, Scene Revisit, and general-metric performance.

\begin{figure*}[t]
  \centering
  \includegraphics[width=\linewidth]{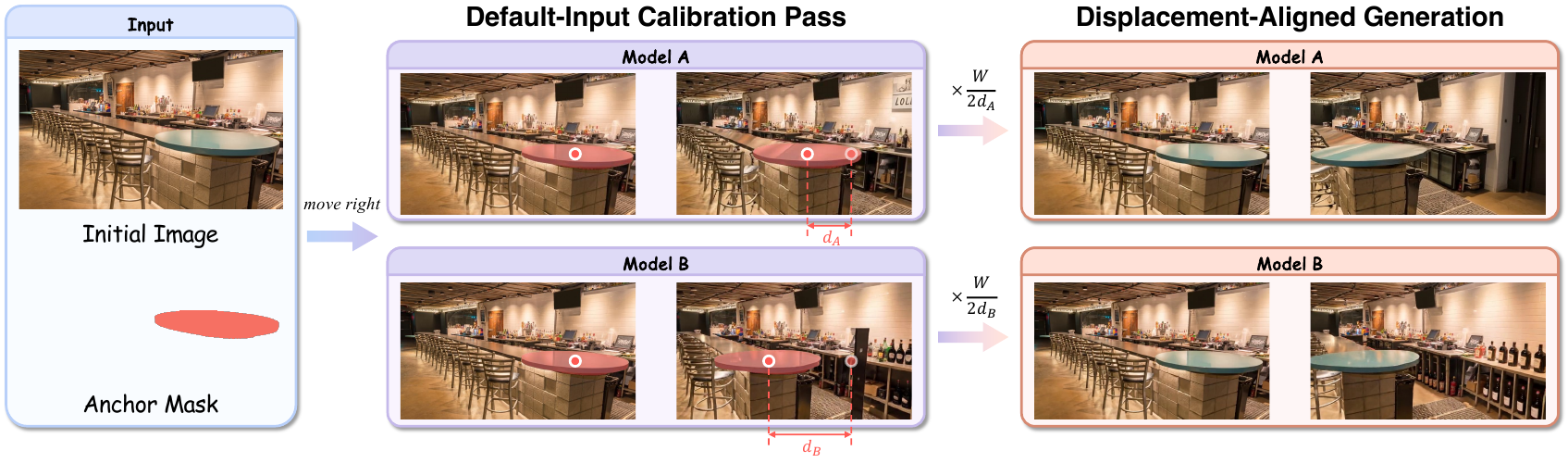}
  \caption{\textbf{Image-space displacement alignment for camera-driven models.}
  Given an initial image and anchor mask for case $c$, \name measures model $m$'s image-space displacement $d_{m,c}$ in a default-input calibration pass and scales its input translations by $k_{m,c}=W/(2d_{m,c})$ to align displacement across camera-driven models.}
  \label{fig:motion-scale-alignment}
\end{figure*}

The scale-invariant translation error in \cref{sec:eval:geometry} does not address this difference.
Its scalar $s$ is fitted after generation only to resolve the coordinate-scale mismatch between the recovered and reference trajectories; it neither changes the input trajectory nor normalizes the image-space displacement in the generated video.
We therefore calibrate input translations before generation to align image-space displacement (\cref{fig:motion-scale-alignment}).

\begin{figure}[t]
  \centering
  \includegraphics[width=\linewidth]{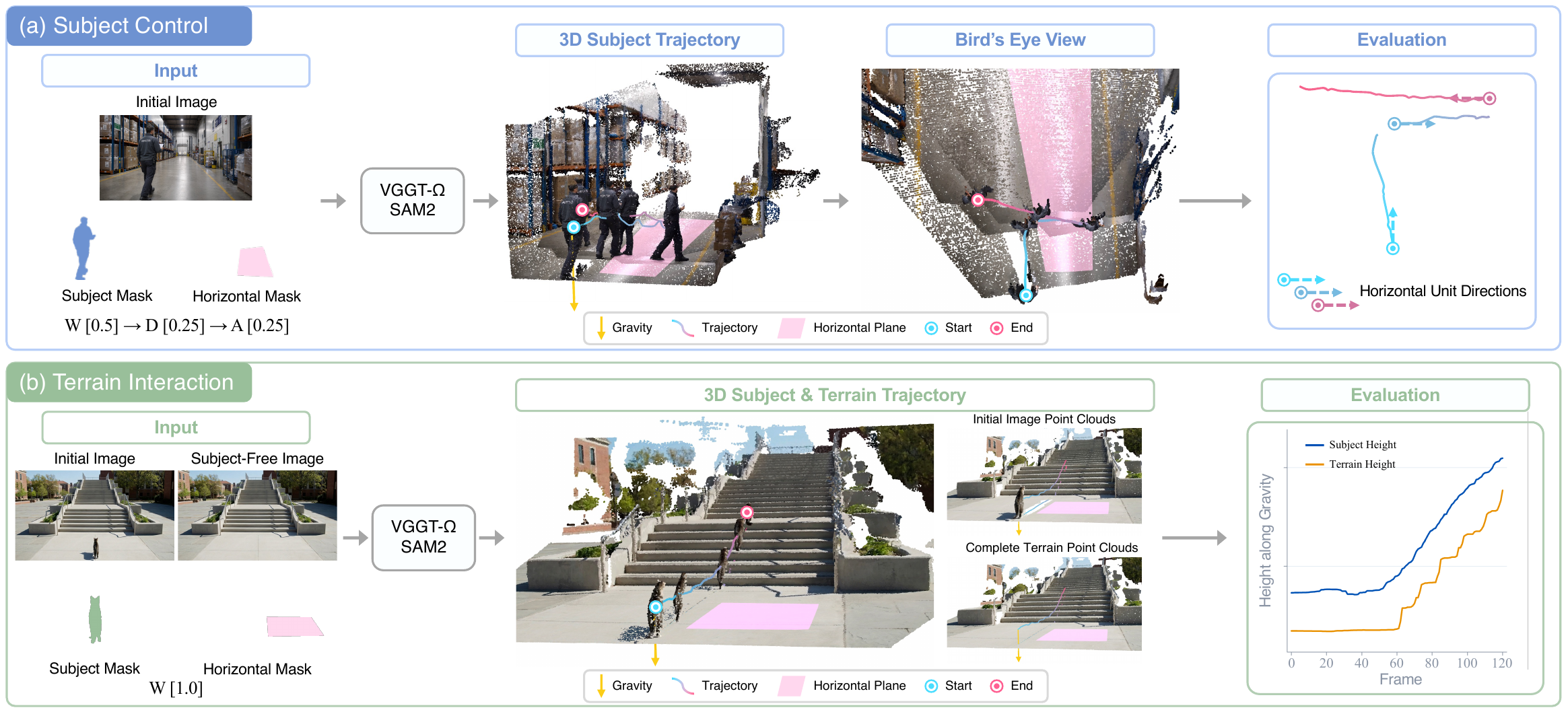}
  \caption{\textbf{Geometry-based evaluation of Subject Control and Terrain Interaction.}
  (a) SAM2 and VGGT-$\Omega$ recover the subject trajectory, while the horizontal-region mask estimates gravity and the horizontal plane used to define the control directions.
  (b) A subject-free image restores the complete terrain geometry, onto which the trajectory is projected along gravity to obtain corresponding terrain trajectory.}
  \label{fig:subject-terrain-evaluation}
\end{figure}

For each model $m$ and case $c$, we estimate a translation calibration factor using the anchor mask provided on the initial frame.
We first generate a calibration video with the model's default input translation magnitude, apply a horizontal camera control, either ``move left'' or ``move right'', and track the anchor through the video using SAM2~\citep{sam2}.
Let $d_{m,c}$ denote the horizontal image-space displacement of the tracked mask center, and let $W$ denote the frame width.
We set the target displacement to $d^*=W/2$ and compute $k_{m,c}=d^*/d_{m,c}=W/(2d_{m,c})$.
For the final generation used in evaluation, we multiply the translation components of the default input trajectory by $k_{m,c}$ while leaving its rotations unchanged.

\subsection{Dynamic-Interaction Track}
\label{sec:eval:dynamic}

\paragraph{\textbf{Subject Control}.}
Subject Control uses the same ordered-control construction as Camera Control, with $a_i\in$ \{W, S, A, D\} applied to a designated subject.
Action-driven models receive native discrete subject actions over the allocated frame ranges.
For language-driven models, we verbalize the ordered controls together with the designated subject and prepend the resulting instruction to the scene description; for example, ``W'' followed by ``A'' becomes ``The [subject] moves forward, then the [subject] moves left. [Scene description].''

As illustrated in \cref{fig:subject-terrain-evaluation}, SAM2~\citep{sam2} tracks the designated subject from a first-frame mask, and VGGT-$\Omega$~\citep{wang2026vggt} lifts the tracked pixels into 3D to recover the subject trajectory.
The horizontal-region mask estimates gravity and the horizontal plane; projecting the initial camera's viewing direction onto this plane defines the forward reference direction, with the other directions derived analogously.

We evaluate the recovered subject trajectory segment by segment.
The action-driven segment boundaries follow the frame ranges assigned to the discrete subject actions, whereas the $N$ language-driven segments are inferred by applying the same change-point procedure as in Camera Control to frame-to-frame subject displacement.
Within each segment, we translate the recovered trajectory so that its first-frame subject position is the origin.
The associated atomic control unit selects a horizontal unit direction $\mathbf{u}_i$, and the reference trajectory is linearly interpolated from the origin to $\mathbf{u}_i$.
We fit a post-generation translation scale $s$ between the recovered and reference trajectories, as in Camera Control, and compute the segment-level translation error $e_{t,i}$.
A segment whose net displacement is below 0.5\% of the reconstructed scene scale or whose motion is misaligned with the prescribed direction receives the maximum error $e_{t,i}=0.5$.
Finally, the error $e_t$ is obtained by frame-count-weighted averaging, and the Subject Control score is $S_{\mathrm{sub}}=100\max(0,1-e_t/0.5)$.

\paragraph{\textbf{Terrain Interaction}.}
Unlike Subject Control, Terrain Interaction uses a single atomic control unit to induce a subject-terrain interaction.
For evaluation, we additionally provide a subject-free terrain image to recover the complete terrain geometry.
We then project the recovered 3D subject trajectory along gravity onto this geometry to obtain the corresponding 3D terrain trajectory.

During evaluation, we first compute the Subject Control score from the horizontal component of the subject trajectory and use it as a gating check; cases that fail this check are considered not to follow the control and receive a Terrain Interaction score of zero.
For cases that pass this check, we extract local extrema and the endpoint from the height of the terrain trajectory as evaluation points.
The Terrain Interaction score is the ratio of the number of evaluation points at which the subject and terrain trajectories exhibit consistent height changes to the total number of evaluation points.

\paragraph{\textbf{Checklist-Based World Reactivity Evaluation}.}
Considering that the remaining four tasks require semantic and causal judgments that cannot be captured by recovered trajectories, we evaluate them with a VLM judge against the case-specific checklist $L=\{\ell_k\}_{k=1}^{K}$ constructed in \cref{sec:design:curation}.
Each checklist covers the initiating condition, the resulting reaction or goal execution progress, and invalid outcomes.
At evaluation time, the VLM receives 10 temporally ordered frames uniformly sampled from the generated video, together with the checklist.
Using a task-specific prompt, the judge returns one binary decision per item; contradicted, missing, ambiguous, off-screen, or otherwise unverifiable evidence is counted as unsatisfied.
The case score is
\begin{equation}
S_{\mathrm{check}}(V,L)=\frac{100}{K}\sum_{k=1}^{K}\mathbb{I}[\ell_k \text{ is satisfied in } V].
\end{equation}

\paragraph{\textbf{Object Interaction}.}
The checklist verifies that contact occurs with the designated object, precedes and causes the reaction, and produces a type-consistent outcome without interpenetration.
It also checks that the direction and extent of the reaction remain consistent with the contact.

\paragraph{\textbf{Social Interaction}.}
The checklist verifies that the controlled motion creates the intended conflict and that at least one visible affected agent makes a timely adjustment attributable to the controlled subject.
Unchanged, delayed, unrelated, or physically implausible responses are counted as failures.

\paragraph{\textbf{Physical Reaction}.}
The checklist verifies the timing and cause of the process, its evolution under the relevant physical regularity, and the preservation of required supports, attachments, contacts, and constraints.
Freezing, premature onset, interpenetration, broken attachments, or unexplained energy are counted as failures.

\paragraph{\textbf{Goal Completion}.}
The checklist separately evaluates correct grounding, intermediate progress, compliance with stated ordering and scene-dependent constraints, and final completion.
Scores credit partial progress and accept alternative executions that reach the desired target under the same observable requirements.

\paragraph{\textbf{General metrics}.}
The dynamic-interaction track separately reports four VBench metrics~\citep{huang2024vbench}: Subject Consistency and Motion Smoothness for feature and temporal consistency, and Aesthetic Quality and Imaging Quality for visual appeal and frame-level quality.
We omit 3D Consistency, Photometric Consistency, and Temporal Flickering because valid motion and state changes disrupt their geometric and optical-flow correspondences.

%% file: sections/5_experiments.tex
\section{Experiments}
\label{sec:experiments}
\input{tables/static_scene_results}

\subsection{Experimental Setup}
\label{sec:experiments:setup}

We evaluate 20 representative video world models: 6 camera-driven, 7 action-driven, and 7 language-driven models.
For each case, we construct the model-facing input in the model's native format using the interface adaptation described in \cref{sec:design} and evaluate the generated video with the protocols in \cref{sec:eval}.
All 20 models are evaluated on the \emph{static-scene track}.
The \emph{dynamic-interaction track} requires observable third-person subject-scene interaction and therefore includes WorldPlay~\citep{worldplay}, LingBot-World~\citep{lingbot}, and all seven language-driven models.
The other five action-driven models either do not support third-person subject control or cannot control a visible third-person subject reliably, and are therefore excluded from the dynamic-interaction track.
Camera-driven models are excluded because their interfaces control only the camera, and Goal Completion is evaluated only on language-driven models.
To avoid compromising model performance, we use each model's default resolution, video length, and other inference settings whenever applicable.
For streaming models with flexible generation lengths, we constrain the output to 100--200 frames to prevent quality drift in excessively long generations from biasing the evaluation.
Some closed-source commercial language-driven systems apply proprietary prompt enhancement before video generation.
Consistent with the functional formulation in \cref{sec:design}, we retain such default preprocessing as part of the native end-to-end pipeline.
Detailed inference settings and task eligibility are provided in \cref{sec:appendix:model-settings}; unsupported tasks are marked with dashes in the result tables.

\subsection{Static-Scene Track Results}
\label{sec:experiments:static}

\Cref{tab:static-scene-results} reports Camera Control and Scene Revisit together with the task-agnostic general metrics.
The task scores diagnose \emph{Control Adherence} and \emph{Spatial Consistency}, whereas the general metrics characterize \emph{Visual Quality}; reporting them separately reveals that these capabilities do not necessarily improve together.

\paragraph{\textbf{Camera Control}.}
Among camera-driven models, the strongest results come from methods that reconstruct 3D priors and reproject them to target views: NeoVerse~\citep{neoverse} scores 97.33 and InSpatio-World~\citep{inspatio} scores 85.94.
ReCamMaster~\citep{recammaster} and FantasyWorld~\citep{fantasyworld}, which encode camera poses as learned tokens or embeddings, obtain much lower Camera Control scores of 38.64 and 18.46 despite competitive General averages of 80.97 and 80.23.
WorldPlay~\citep{worldplay} is the strongest action-driven model at 92.74.
Language-driven models are less precise, with Hailuo 2.3~\citep{hailuo} achieving the strongest score of 63.29, consistent with the difficulty of expressing ordered, complex viewpoint changes through natural-language instructions.

\paragraph{\textbf{Scene Revisit}.}
NeoVerse and InSpatio-World both achieve 1.000 Revisit Success and lead the camera-driven group with Scene Revisit scores of 89.25 and 85.90, respectively.
Among action-driven models, WorldPlay performs best with 0.790 Revisit Success and a score of 72.51, while Hailuo 2.3 leads the language-driven group with 0.505 and 48.70.
The remaining gap reflects failures either to return to the initial viewpoint or to recover its geometry, appearance, and content after the round trip.

\subsection{Effect of the Input Translation Multiplier}
\label{sec:experiments:scale-ablation}

\begin{wrapfigure}{l}{0.3\linewidth}
  \centering
  \vspace{-7mm}
  \includegraphics[width=\linewidth]{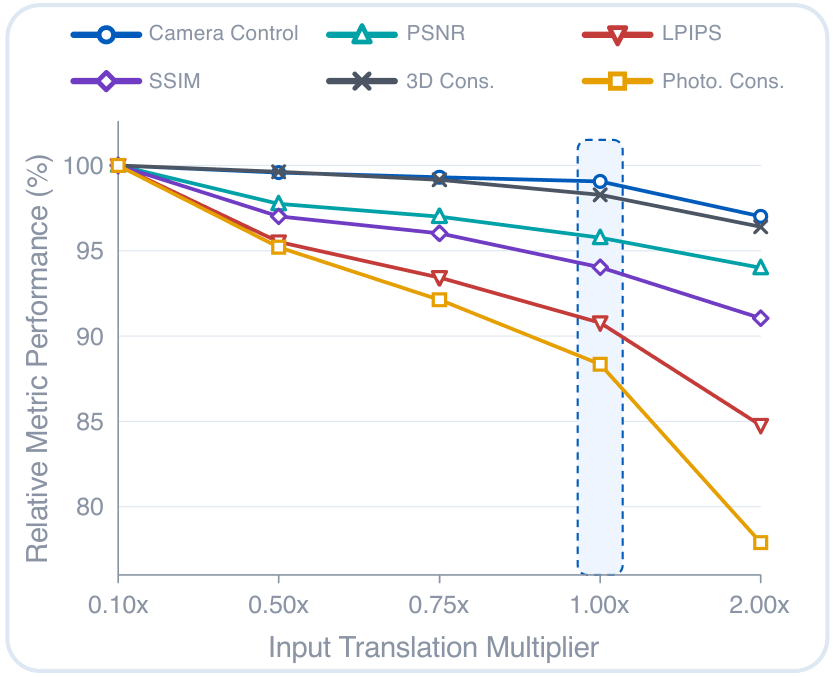}
  \vspace{-6.9mm}
  \caption{\textbf{Effect of the input translation multiplier on NeoVerse.}
  }
  \vspace{-10mm}
  \label{fig:scale-ablation}
\end{wrapfigure}

We evaluate NeoVerse~\citep{neoverse} under the same camera controls while varying only the translation multiplier of its input $\mathrm{SE}(3)$ trajectory.
As the multiplier increases from $0.10\times$ to $2.00\times$, the Camera Control score decreases from 98.25 to 95.32, the Scene Revisit score from 90.98 to 88.37, and the General average from 80.42 to 75.05; all five general metrics decline, with Photometric Consistency falling from 80.17 to 62.45.
The results confirm that larger image-space displacements make both controlled generation and scene preservation more difficult.
This ablation therefore motivates the pre-generation image-space displacement alignment in \cref{sec:eval:scale-alignment}, which is applied to all reported camera-driven results.
\input{tables/camera_scale_alignment_ablation}

\subsection{Dynamic-Interaction Track Results}
\label{sec:experiments:interactive}

\input{tables/interactive_results}

\Cref{tab:interactive-results} reveals whether a model follows subject-motion control and whether it can react correctly.

\paragraph{\textbf{Subject Control}.}
The direct action interfaces provide more precise subject control: LingBot-World~\citep{lingbot} scores 55.47 and WorldPlay~\citep{worldplay} scores 49.75, compared with the best language-driven score of 37.28 from Veo 3.1~\citep{veo}.
Even the action-driven results remain far from saturated, with failures often converting the requested subject motion into camera motion or leaving the scene static.

\paragraph{\textbf{Terrain Interaction}.}
Vidu Q3~\citep{vidu} and Hailuo 2.3~\citep{hailuo} lead with 64.39 and 61.57, whereas the best action-driven score is 27.49.
For action-driven models, the large drop from Subject Control to Terrain Interaction shows that horizontal control adherence does not guarantee vertical terrain adaptation.

\paragraph{\textbf{Object Interaction}.}
Veo 3.1 and Vidu Q3 lead with 75.96 and 71.59, while the best action-driven score is 33.75.
Common failures leave the contacted object unchanged or allow the subject to pass through it.

\paragraph{\textbf{Social Interaction}.}
Veo 3.1 achieves the highest score of 85.10, followed by Vidu Q3 at 81.91; the best action-driven score is 60.37.
Failures typically leave nearby agents unresponsive or allow the controlled subject to move through them without avoidance or yielding.

\paragraph{\textbf{Physical Reaction}.}
Hailuo 2.3 leads with 63.84, followed by Veo 3.1 and Vidu Q3 at 61.76 and 61.23; the best action-driven score is 33.43.
Action-driven generations often execute subject control while leaving unstable or contacted objects unchanged, exposing the gap between explicit control and inherent reactivity.

\paragraph{\textbf{Goal Completion}.}
HappyHorse 1.0~\citep{happyhorse} and Veo 3.1 achieve the strongest results at 85.33 and 85.30.
Kling 2.5~\citep{kling} scores only 48.25 despite having the highest General average among language-driven models, showing that visual quality does not guarantee grounded goal-directed execution.

\subsection{Cross-Task Diagnostic Analysis}
\label{sec:experiments:diagnostic}

\begin{figure}[t]
  \centering
  \includegraphics[width=\linewidth]{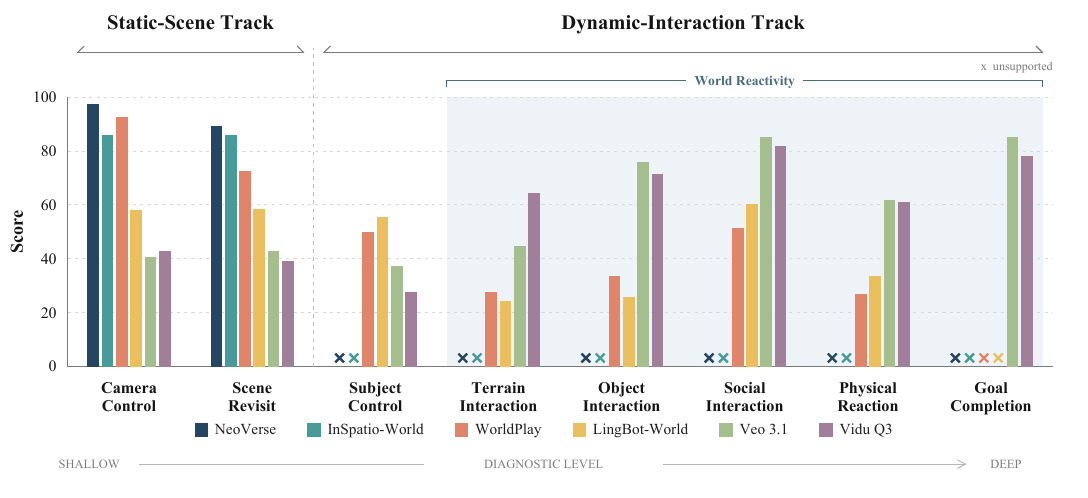}
  \caption{\textbf{Task-level performance across evaluation tracks.}
  Two models per interface are compared across all eight tasks, ordered from static-scene diagnostics through Subject Control to World Reactivity; crosses mark unsupported tasks.}
  \label{fig:task-result}
\end{figure}

\Cref{fig:task-result} consolidates the capability split across the three interfaces.
Camera-driven models provide the strongest camera control and scene revisiting but do not support dynamic interaction.
Action-driven models control designated subjects more precisely, yet this advantage does not consistently transfer to the scene-conditioned reactions induced by those controls.
Language-driven models perform better on interaction and goal-directed tasks but follow composed camera and subject controls less faithfully.
No model combines broad coverage with consistently strong performance, leaving current interfaces complementary but incomplete.

The split is also obscured by general visual metrics.
For example, the language-driven General averages occupy a narrow range of 79.64--81.04, while their Task averages range from 39.85 to 65.02.
Likewise, ReCamMaster and FantasyWorld retain strong General averages despite weak Camera Control scores.
These gaps show that the four diagnostic levels capture distinct capabilities.
In particular, strong \emph{Visual Quality} or \emph{Control Adherence} does not guarantee \emph{World Reactivity}, motivating separate reporting of the four levels.

\subsection{Human Alignment of Checklist Evaluation}
\label{sec:experiments:human}

\input{tables/human_alignment}

We validate the VLM judge on the four tasks evaluated using checklists.
The validation set contains 800 evaluation instances and 5{,}793 checklist items, with 200 instances per task.
Three human annotators independently label each item from the same 10 temporally ordered frames shown to the VLM, and the majority vote defines the binary reference label.
For each instance, the human and VLM scores are computed as the respective fractions of satisfied checklist items.
\Cref{tab:human-alignment} reports Spearman's $\rho$ and PLCC for each task separately and across all 800 evaluation instances combined.
Across all 800 instances, Spearman's $\rho$ is 0.8614 and PLCC is 0.8583, showing strong agreement between the VLM judge and human evaluation.
\input{sections/5_additional_experiments}

%% file: tables/static_scene_results.tex
\begin{table*}[t]
\centering
\small
\caption{\textbf{Static-scene track evaluation.}
\textbf{Task} averages the Camera Control and Scene Revisit scores, \textbf{General} averages the five general metrics, and \textbf{Overall} averages Task and General scores.
Down $\downarrow$ and up $\uparrow$ arrows indicate that lower and higher values are better, respectively.
The best and second-best results per paradigm are bold and underlined.}
\label{tab:static-scene-results}
\setlength{\heavyrulewidth}{0.112em}
\setlength{\lightrulewidth}{0.070em}
\setlength{\cmidrulewidth}{0.042em}
\resizebox{\linewidth}{!}{%
\setlength{\tabcolsep}{3pt}
\renewcommand{\arraystretch}{1.03}
\begin{tabular*}{1.40\linewidth}{@{\extracolsep{\fill}}l*{16}{c}@{}}
\toprule
\multirow[c]{2}{*}[-2pt]{\textbf{Model}} & \multicolumn{3}{c}{\textbf{Camera Control}} & \multicolumn{5}{c}{\textbf{Scene Revisit}} & \multirow[c]{2}{*}[-2pt]{\shortstack{\textbf{3D}\\\textbf{Cons.}}} & \multirow[c]{2}{*}[-2pt]{\shortstack{\textbf{Photo.}\\\textbf{Cons.}}} & \multirow[c]{2}{*}[-2pt]{\shortstack{\textbf{Temp.}\\\textbf{Flick.}}} & \multirow[c]{2}{*}[-2pt]{\shortstack{\textbf{Aesth.}\\\textbf{Quality}}} & \multirow[c]{2}{*}[-2pt]{\shortstack{\textbf{Imag.}\\\textbf{Quality}}} & \multicolumn{3}{c}{\textbf{Average}} \\
\cmidrule(lr){2-4}\cmidrule(lr){5-9}\cmidrule(lr){15-17}
& \textbf{T. Err.}$\downarrow$ & \textbf{R. Err.}$\downarrow$ & \textbf{Score}$\uparrow$ & \textbf{Success}$\uparrow$ & \textbf{PSNR}$\uparrow$ & \textbf{LPIPS}$\downarrow$ & \textbf{SSIM}$\uparrow$ & \textbf{Score}$\uparrow$ & & & & & & \textbf{Task} & \textbf{General} & \textbf{Overall} \\
\midrule
\rowcolor{gray!15}
\multicolumn{17}{@{}l}{\textit{Camera-driven}} \\
TrajectoryCrafter~\citep{trajectorycrafter} & 0.14 & \underline{1.01} & 80.32 & \textbf{1.000} & 19.08 & 0.240 & 0.545 & 83.03 & 96.98 & 62.83 & 92.56 & 52.17 & 67.03 & 81.68 & 74.31 & 78.00 \\
ReCamMaster~\citep{recammaster} & 0.47 & 3.91 & 38.64 & 0.815 & 16.02 & 0.353 & 0.415 & 68.01 & \textbf{99.33} & \textbf{81.97} & \underline{95.52} & \underline{54.58} & 73.45 & 53.33 & \textbf{80.97} & 67.15 \\
Voyager~\citep{voyager} & 0.31 & 4.29 & 56.19 & \underline{0.995} & 16.88 & 0.388 & 0.466 & 76.27 & 89.54 & 27.43 & 93.04 & 53.59 & 63.69 & 66.23 & 65.46 & 65.84 \\
FantasyWorld~\citep{fantasyworld} & 1.04 & 7.82 & 18.46 & 0.560 & 15.41 & 0.323 & 0.374 & 55.79 & \underline{98.57} & \underline{75.60} & \textbf{95.91} & \textbf{57.11} & \textbf{73.98} & 37.12 & \underline{80.23} & 58.68 \\
NeoVerse~\citep{neoverse} & \textbf{0.01} & \textbf{0.60} & \textbf{97.33} & \textbf{1.000} & \textbf{21.72} & \textbf{0.141} & \textbf{0.662} & \textbf{89.25} & 98.19 & 70.83 & 93.53 & 52.72 & 72.19 & \textbf{93.29} & 77.49 & \textbf{85.39} \\
InSpatio-World (1.3B)~\citep{inspatio} & \underline{0.09} & 1.24 & \underline{85.94} & \textbf{1.000} & \underline{20.79} & \underline{0.224} & \underline{0.606} & \underline{85.90} & 98.44 & 64.88 & 93.68 & 53.78 & \underline{73.60} & \underline{85.92} & 76.88 & \underline{81.40} \\
\midrule
\rowcolor{gray!15}
\multicolumn{17}{@{}l}{\textit{Action-driven}} \\
Hunyuan-GameCraft~\citep{gamecraft} & 0.09 & 11.95 & 41.33 & 0.385 & 13.36 & 0.542 & 0.367 & 41.77 & 93.61 & 53.93 & 93.91 & 54.97 & 71.69 & 41.55 & 73.62 & 57.59 \\
Astra~\citep{astra} & 0.31 & 7.92 & 32.59 & 0.365 & 12.20 & 0.601 & 0.309 & 38.15 & 96.46 & 78.41 & \underline{96.38} & 51.71 & 71.78 & 35.37 & 78.95 & 57.16 \\
WorldPlay~\citep{worldplay} & \textbf{0.04} & \textbf{1.04} & \textbf{92.74} & \underline{0.790} & \textbf{18.42} & \textbf{0.271} & \textbf{0.531} & \textbf{72.51} & 98.69 & 81.06 & 95.90 & 53.92 & 73.31 & \textbf{82.63} & 80.58 & \textbf{81.61} \\
Yume 1.5~\citep{yume15} & 0.12 & 3.23 & 75.67 & 0.255 & 12.37 & 0.615 & 0.359 & 32.45 & 98.44 & 67.53 & 95.18 & 53.42 & \underline{75.28} & 54.06 & 77.97 & 66.02 \\
LingBot-World~\citep{lingbot} & 0.11 & 6.22 & 58.19 & 0.605 & \underline{15.58} & \underline{0.341} & \underline{0.406} & 58.35 & \underline{99.59} & \underline{81.52} & \textbf{96.53} & \textbf{59.87} & 75.00 & 58.27 & \underline{82.50} & 70.39 \\
Infinite-World~\citep{infiniteworld} & 0.17 & \underline{2.35} & 71.70 & 0.645 & 14.31 & 0.388 & 0.345 & 57.34 & \textbf{99.91} & \textbf{92.09} & 95.99 & \underline{56.27} & \textbf{76.41} & 64.52 & \textbf{84.13} & \underline{74.33} \\
Matrix-Game 3.0~\citep{matrixgame3} & \underline{0.06} & 4.36 & \underline{76.36} & \textbf{0.860} & 13.82 & 0.485 & 0.372 & \underline{64.25} & 95.59 & 30.14 & 93.40 & 48.80 & 73.39 & \underline{70.30} & 68.26 & 69.28 \\
\midrule
\rowcolor{gray!15}
\multicolumn{17}{@{}l}{\textit{Language-driven}} \\
Kling 2.5~\citep{kling} & 0.19 & 6.90 & 50.18 & 0.325 & \textbf{13.54} & 0.532 & \underline{0.381} & 38.81 & \textbf{99.87} & \textbf{87.87} & \textbf{97.56} & \underline{56.28} & 75.70 & 44.50 & \textbf{83.46} & 63.98 \\
Veo 3.1~\citep{veo} & 0.27 & 8.22 & 40.83 & 0.447 & 12.30 & 0.594 & 0.345 & 43.05 & 99.06 & 73.61 & \underline{95.43} & 55.49 & 76.57 & 41.94 & 80.03 & 60.99 \\
Hailuo 2.3~\citep{hailuo} & \textbf{0.15} & \textbf{5.20} & \textbf{63.29} & \textbf{0.505} & \underline{13.52} & \textbf{0.519} & \textbf{0.387} & \textbf{48.70} & 99.38 & \underline{79.88} & 95.33 & \textbf{56.30} & 75.52 & \textbf{55.99} & \underline{81.28} & \textbf{68.64} \\
Wan 2.6 I2V~\citep{wan} & \underline{0.17} & 6.12 & 57.72 & \underline{0.495} & 13.28 & \underline{0.527} & 0.353 & \underline{47.32} & 99.49 & 69.42 & 94.13 & 53.93 & \textbf{77.53} & \underline{52.52} & 78.90 & \underline{65.71} \\
Seedance 1.5~\citep{seedance} & 0.20 & 7.69 & 49.18 & 0.375 & 12.28 & 0.602 & 0.342 & 39.23 & 97.31 & 56.60 & 94.81 & 54.45 & 74.88 & 44.21 & 75.61 & 59.91 \\
Vidu Q3~\citep{vidu} & 0.24 & 8.47 & 42.75 & 0.360 & 12.65 & 0.586 & 0.351 & 39.05 & 99.43 & 71.28 & 95.01 & 55.38 & 76.73 & 40.90 & 79.57 & 60.24 \\
HappyHorse 1.0~\citep{happyhorse} & 0.18 & \underline{5.62} & \underline{58.29} & 0.420 & 12.48 & 0.564 & 0.352 & 42.45 & \underline{99.62} & 74.17 & 95.02 & 55.56 & \underline{77.00} & 50.37 & 80.27 & 65.32 \\
\bottomrule
\end{tabular*}%
}
\end{table*}

%% file: tables/camera_scale_alignment_ablation.tex
\begin{table*}[t]
\centering
\small
\caption{\textbf{Effect of the input translation multiplier on NeoVerse.}
The multiplier is applied only to the translation components of the input $\mathrm{SE}(3)$ trajectory; rotations remain unchanged.
Metrics and the Task, General, and Overall aggregates follow \cref{tab:static-scene-results}.}
\label{tab:camera-scale-alignment-ablation}
\setlength{\heavyrulewidth}{0.112em}
\setlength{\lightrulewidth}{0.070em}
\setlength{\cmidrulewidth}{0.042em}
\resizebox{\linewidth}{!}{%
\setlength{\tabcolsep}{3pt}
\renewcommand{\arraystretch}{1.03}
\begin{tabular*}{1.40\linewidth}{@{\extracolsep{\fill}}c*{16}{c}@{}}
\toprule
\multirow[c]{2}{*}[-2pt]{\shortstack{\textbf{Translation}\\\textbf{Multiplier}}} & \multicolumn{3}{c}{\textbf{Camera Control}} & \multicolumn{5}{c}{\textbf{Scene Revisit}} & \multirow[c]{2}{*}[-2pt]{\shortstack{\textbf{3D}\\\textbf{Cons.}}} & \multirow[c]{2}{*}[-2pt]{\shortstack{\textbf{Photo.}\\\textbf{Cons.}}} & \multirow[c]{2}{*}[-2pt]{\shortstack{\textbf{Temp.}\\\textbf{Flick.}}} & \multirow[c]{2}{*}[-2pt]{\shortstack{\textbf{Aesth.}\\\textbf{Quality}}} & \multirow[c]{2}{*}[-2pt]{\shortstack{\textbf{Imag.}\\\textbf{Quality}}} & \multicolumn{3}{c}{\textbf{Average}} \\
\cmidrule(lr){2-4}\cmidrule(lr){5-9}\cmidrule(lr){15-17}
& \textbf{T. Err.}$\downarrow$ & \textbf{R. Err.}$\downarrow$ & \textbf{Score}$\uparrow$ & \textbf{Success}$\uparrow$ & \textbf{PSNR}$\uparrow$ & \textbf{LPIPS}$\downarrow$ & \textbf{SSIM}$\uparrow$ & \textbf{Score}$\uparrow$ & & & & & & \textbf{Task} & \textbf{General} & \textbf{Overall} \\
\midrule
0.10$\times$ & 0.002 & 0.43 & 98.25 & 1.000 & 22.68 & 0.128 & 0.704 & 90.98 & 99.91 & 80.17 & 95.41 & 54.06 & 72.53 & 94.62 & 80.42 & 87.52 \\
0.50$\times$ & 0.003 & 0.51 & 97.84 & 1.000 & 22.17 & 0.134 & 0.683 & 90.11 & 99.54 & 76.33 & 94.21 & 53.39 & 72.47 & 93.98 & 79.19 & 86.59 \\
0.75$\times$ & 0.004 & 0.56 & 97.57 & 1.000 & 22.00 & 0.137 & 0.676 & 89.80 & 99.07 & 73.86 & 93.81 & 53.03 & 72.30 & 93.69 & 78.41 & 86.05 \\
1.00$\times$ & 0.005 & 0.60 & 97.33 & 1.000 & 21.72 & 0.141 & 0.662 & 89.25 & 98.19 & 70.83 & 93.53 & 52.72 & 72.19 & 93.29 & 77.49 & 85.39 \\
2.00$\times$ & 0.015 & 0.91 & 95.32 & 1.000 & 21.32 & 0.151 & 0.641 & 88.37 & 96.32 & 62.45 & 92.93 & 52.00 & 71.54 & 91.85 & 75.05 & 83.45 \\
\bottomrule
\end{tabular*}%
}
\end{table*}

%% file: tables/interactive_results.tex
\begin{table*}[t]
\centering
\small
\caption{\textbf{Dynamic-interaction track evaluation.}
Subject Control and five \emph{World Reactivity} tasks are reported for compatible action- and language-driven models.
\textbf{Task} averages the supported task scores, \textbf{General} averages the four general metrics, and \textbf{Overall} averages Task and General scores.
Down $\downarrow$ and up $\uparrow$ arrows indicate that lower and higher values are better, respectively.
Dashes mark unsupported tasks; the best and second-best results per paradigm are bold and underlined.}
\vspace{-3mm}
\label{tab:interactive-results}
\setlength{\heavyrulewidth}{0.096em}
\setlength{\lightrulewidth}{0.060em}
\setlength{\cmidrulewidth}{0.036em}
\resizebox{\linewidth}{!}{%
\setlength{\tabcolsep}{3pt}
\renewcommand{\arraystretch}{1.03}
\begin{tabular*}{1.20\linewidth}{@{\extracolsep{\fill}}l*{13}{c}@{}}
\toprule
\multirow[c]{2}{*}[-2pt]{\textbf{Model}} & \multirow[c]{2}{*}[-2pt]{\shortstack{\textbf{Subject}\\\textbf{Control}}} & \multirow[c]{2}{*}[-2pt]{\shortstack{\textbf{Terrain}\\\textbf{Inter.}}} & \multirow[c]{2}{*}[-2pt]{\shortstack{\textbf{Object}\\\textbf{Inter.}}} & \multirow[c]{2}{*}[-2pt]{\shortstack{\textbf{Social}\\\textbf{Inter.}}} & \multirow[c]{2}{*}[-2pt]{\shortstack{\textbf{Physical}\\\textbf{Reaction}}} & \multirow[c]{2}{*}[-2pt]{\shortstack{\textbf{Goal}\\\textbf{Completion}}} & \multirow[c]{2}{*}[-2pt]{\shortstack{\textbf{Subject}\\\textbf{Cons.}}} & \multirow[c]{2}{*}[-2pt]{\shortstack{\textbf{Motion}\\\textbf{Smooth.}}} & \multirow[c]{2}{*}[-2pt]{\shortstack{\textbf{Aesth.}\\\textbf{Quality}}} & \multirow[c]{2}{*}[-2pt]{\shortstack{\textbf{Imag.}\\\textbf{Quality}}} & \multicolumn{3}{c}{\textbf{Average}} \\
\cmidrule(lr){12-14}
& & & & & & & & & & & \textbf{Task} & \textbf{General} & \textbf{Overall} \\
\midrule
\rowcolor{gray!15}
\multicolumn{14}{@{}l}{\textit{Action-driven}} \\
WorldPlay~\citep{worldplay} & \underline{49.75} & \textbf{27.49} & \textbf{33.75} & \underline{51.40} & \underline{26.91} & -- & \underline{88.06} & \underline{98.09} & \underline{54.19} & \underline{68.76} & \underline{37.86} & \underline{77.28} & \underline{57.57} \\
LingBot-World~\citep{lingbot} & \textbf{55.47} & \underline{24.33} & \underline{25.94} & \textbf{60.37} & \textbf{33.43} & -- & \textbf{94.98} & \textbf{98.89} & \textbf{60.86} & \textbf{71.69} & \textbf{39.91} & \textbf{81.61} & \textbf{60.76} \\
\midrule
\rowcolor{gray!15}
\multicolumn{14}{@{}l}{\textit{Language-driven}} \\
Kling 2.5~\citep{kling} & 28.40 & 35.95 & 27.70 & 66.80 & 31.99 & 48.25 & \textbf{96.00} & \textbf{99.48} & 56.86 & 71.83 & 39.85 & \textbf{81.04} & 60.45 \\
Veo 3.1~\citep{veo} & \textbf{37.28} & 44.71 & \textbf{75.96} & \textbf{85.10} & \underline{61.76} & \underline{85.30} & 92.07 & 99.12 & \textbf{58.22} & 72.65 & \textbf{65.02} & 80.52 & \textbf{72.77} \\
Hailuo 2.3~\citep{hailuo} & \underline{36.49} & \underline{61.57} & 67.01 & 72.45 & \textbf{63.84} & 78.86 & 93.25 & \underline{99.31} & \underline{57.74} & 72.47 & 63.37 & 80.69 & 72.03 \\
Wan 2.6 I2V~\citep{wan} & 29.02 & 49.21 & 44.56 & 66.40 & 48.15 & 76.39 & \underline{94.46} & 98.15 & 56.30 & \textbf{74.72} & 52.29 & \underline{80.91} & 66.60 \\
Seedance 1.5~\citep{seedance} & 32.51 & 53.83 & 37.91 & 72.09 & 47.60 & 76.24 & 92.14 & 98.87 & 56.72 & 70.84 & 53.36 & 79.64 & 66.50 \\
Vidu Q3~\citep{vidu} & 27.67 & \textbf{64.39} & \underline{71.59} & \underline{81.91} & 61.23 & 78.26 & 92.78 & 98.76 & 57.24 & \underline{73.25} & \underline{64.18} & 80.51 & \underline{72.35} \\
HappyHorse 1.0~\citep{happyhorse} & 33.11 & 56.30 & 65.70 & 76.17 & 47.01 & \textbf{85.33} & 92.69 & 98.85 & 57.37 & 73.23 & 60.60 & 80.54 & 70.57 \\
\bottomrule
\vspace{-10mm}
\end{tabular*}%
}
\end{table*}

%% file: tables/human_alignment.tex
\begin{wraptable}{l}{0.4\linewidth}
\vspace{-4mm}
\centering
\small
\caption{\textbf{Human alignment of checklist evaluation.}
Spearman's $\rho$ and PLCC (Pearson correlation) compare human and VLM checklist-satisfaction scores per task and across all 800 instances.}
\vspace{-0.5\baselineskip}
\label{tab:human-alignment}
\resizebox{\linewidth}{!}{%
\setlength{\tabcolsep}{3pt}
\renewcommand{\arraystretch}{1.03}
\begin{tabular}{@{}lccc@{}}
\toprule
Task & \shortstack{Checklist\\Items} & \shortstack{Spearman\\$\rho$} & PLCC \\
\midrule
Goal Completion & 1{,}303 & 0.8960 & 0.9017 \\
Physical Reaction & 1{,}425 & 0.8838 & 0.8750 \\
Object Interaction & 1{,}527 & 0.8251 & 0.8541 \\
Social Interaction & 1{,}538 & 0.7019 & 0.7103 \\
\midrule
\textbf{Overall} & \textbf{5{,}793} & \textbf{0.8614} & \textbf{0.8583} \\
\bottomrule
\end{tabular}%
}
\vspace{-5mm}
\end{wraptable}

%% file: sections/5_additional_experiments.tex
\subsection{Backend Stability with DA3 Reconstruction}
\label{sec:experiments:da3-backend}

To assess sensitivity to the reconstruction backend, we rerun the geometry-based metrics with Depth Anything 3 (DA3)~\citep{da3} while keeping the benchmark inputs and model outputs fixed.
\Cref{tab:static-scene-results-da3,tab:interactive-results-da3} report the corresponding results.
Scores for the four tasks evaluated using checklists remain unchanged.

On the static-scene track, the mean absolute relative change in Overall score is 3.09\% across all 20 models: 0.44\% for camera-driven, 3.08\% for action-driven, and 5.36\% for language-driven models.
Most variation is concentrated in Camera Control, while Scene Revisit and the general metrics change little.
NeoVerse, WorldPlay, and Hailuo 2.3 remain the leading models in their respective groups; the camera- and action-driven rankings are fully preserved, with only closely matched language-driven models exchanging positions.

On the dynamic-interaction track, DA3 affects only Subject Control, Terrain Interaction, and their geometry-dependent aggregates.
The mean absolute relative change in Overall score is 0.57\%, the maximum change is 1.16\%, and all within-paradigm rankings are preserved.
Together, these small changes and stable rankings show that the main model comparisons do not depend on a particular reconstruction backend.

\input{tables/static_scene_results_da3}
\input{tables/interactive_results_da3}

%% file: tables/static_scene_results_da3.tex
\begin{table*}[t]
\centering
\small
\caption{\textbf{Static-scene track evaluation with DA3.}
We recompute the static-scene metrics with DA3 while keeping benchmark inputs and model outputs fixed; aggregation and highlights follow \cref{tab:static-scene-results}.}
\label{tab:static-scene-results-da3}
\setlength{\heavyrulewidth}{0.112em}
\setlength{\lightrulewidth}{0.070em}
\setlength{\cmidrulewidth}{0.042em}
\resizebox{\linewidth}{!}{%
\setlength{\tabcolsep}{3pt}
\renewcommand{\arraystretch}{1.03}
\begin{tabular*}{1.40\linewidth}{@{\extracolsep{\fill}}l*{16}{c}@{}}
\toprule
\multirow[c]{2}{*}[-2pt]{\textbf{Model}} & \multicolumn{3}{c}{\textbf{Camera Control}} & \multicolumn{5}{c}{\textbf{Scene Revisit}} & \multirow[c]{2}{*}[-2pt]{\shortstack{\textbf{3D}\\\textbf{Cons.}}} & \multirow[c]{2}{*}[-2pt]{\shortstack{\textbf{Photo.}\\\textbf{Cons.}}} & \multirow[c]{2}{*}[-2pt]{\shortstack{\textbf{Temp.}\\\textbf{Flick.}}} & \multirow[c]{2}{*}[-2pt]{\shortstack{\textbf{Aesth.}\\\textbf{Quality}}} & \multirow[c]{2}{*}[-2pt]{\shortstack{\textbf{Imag.}\\\textbf{Quality}}} & \multicolumn{3}{c}{\textbf{Average}} \\
\cmidrule(lr){2-4}\cmidrule(lr){5-9}\cmidrule(lr){15-17}
& \textbf{T. Err.}$\downarrow$ & \textbf{R. Err.}$\downarrow$ & \textbf{Score}$\uparrow$ & \textbf{Success}$\uparrow$ & \textbf{PSNR}$\uparrow$ & \textbf{LPIPS}$\downarrow$ & \textbf{SSIM}$\uparrow$ & \textbf{Score}$\uparrow$ & & & & & & \textbf{Task} & \textbf{General} & \textbf{Overall} \\
\midrule
\rowcolor{gray!15}
\multicolumn{17}{@{}l}{\textit{Camera-driven}} \\
TrajectoryCrafter~\citep{trajectorycrafter} & 0.13 & \underline{1.10} & 81.97 & \textbf{1.000} & 19.06 & 0.240 & 0.544 & 82.99 & 95.63 & 62.83 & 92.56 & 52.18 & 67.03 & 82.48 & 74.05 & 78.27 \\
ReCamMaster~\citep{recammaster} & 0.47 & 4.01 & 38.19 & 0.815 & 16.03 & 0.352 & 0.415 & 68.04 & \textbf{99.18} & \textbf{81.97} & \underline{95.51} & \underline{54.58} & 73.45 & 53.12 & \textbf{80.94} & 67.03 \\
Voyager~\citep{voyager} & 0.30 & 4.04 & 60.03 & \underline{0.995} & 16.89 & 0.386 & 0.467 & 76.33 & 89.81 & 27.44 & 93.04 & 53.59 & 63.69 & 68.18 & 65.51 & 66.85 \\
FantasyWorld~\citep{fantasyworld} & 1.05 & 7.76 & 18.24 & 0.555 & 15.39 & 0.323 & 0.373 & 55.51 & \underline{98.79} & \underline{75.60} & \textbf{95.90} & \textbf{57.11} & \textbf{73.98} & 36.87 & \underline{80.28} & 58.58 \\
NeoVerse~\citep{neoverse} & \textbf{0.01} & \textbf{0.72} & \textbf{96.93} & \textbf{1.000} & \textbf{21.70} & \textbf{0.141} & \textbf{0.661} & \textbf{89.22} & 97.53 & 70.83 & 93.53 & 52.72 & 72.19 & \textbf{93.07} & 77.36 & \textbf{85.22} \\
InSpatio-World (1.3B)~\citep{inspatio} & \underline{0.08} & 1.21 & \underline{86.91} & \textbf{1.000} & \underline{20.77} & \underline{0.225} & \underline{0.605} & \underline{85.84} & 97.79 & 64.88 & 93.68 & 53.77 & \underline{73.60} & \underline{86.38} & 76.74 & \underline{81.56} \\
\midrule
\rowcolor{gray!15}
\multicolumn{17}{@{}l}{\textit{Action-driven}} \\
Hunyuan-GameCraft~\citep{gamecraft} & \underline{0.13} & 11.92 & 39.17 & 0.375 & 13.37 & 0.542 & 0.369 & 41.26 & 94.00 & 53.93 & 93.91 & 54.98 & 71.70 & 40.21 & 73.70 & 56.96 \\
Astra~\citep{astra} & 0.71 & 7.97 & 15.97 & 0.360 & 12.18 & 0.600 & 0.309 & 37.89 & 97.09 & 78.42 & \underline{96.37} & 51.71 & 71.78 & 26.93 & 79.07 & 53.00 \\
WorldPlay~\citep{worldplay} & \textbf{0.08} & \textbf{1.06} & \textbf{88.31} & \underline{0.780} & \textbf{18.69} & \textbf{0.269} & \textbf{0.544} & \textbf{72.52} & 98.55 & 81.06 & 95.90 & 53.92 & 73.31 & \textbf{80.41} & 80.55 & \textbf{80.48} \\
Yume 1.5~\citep{yume15} & 0.32 & 3.14 & 59.69 & 0.270 & 12.37 & 0.614 & 0.359 & 33.40 & 97.74 & 67.53 & 95.18 & 53.42 & \underline{75.27} & 46.55 & 77.83 & 62.19 \\
LingBot-World~\citep{lingbot} & 0.19 & 6.28 & 54.43 & 0.620 & \underline{15.63} & \underline{0.340} & \underline{0.409} & 59.17 & \underline{99.42} & \underline{81.52} & \textbf{96.53} & \textbf{59.87} & 75.01 & 56.80 & \underline{82.47} & 69.64 \\
Infinite-World~\citep{infiniteworld} & 0.25 & \underline{2.38} & 64.49 & 0.655 & 14.30 & 0.389 & 0.345 & 57.76 & \textbf{99.87} & \textbf{92.09} & 95.99 & \underline{56.27} & \textbf{76.40} & 61.12 & \textbf{84.12} & \underline{72.62} \\
Matrix-Game 3.0~\citep{matrixgame3} & 0.17 & 4.27 & \underline{68.93} & \textbf{0.860} & 13.83 & 0.486 & 0.372 & \underline{64.23} & 95.61 & 30.14 & 93.39 & 48.80 & 73.39 & \underline{66.58} & 68.27 & 67.43 \\
\midrule
\rowcolor{gray!15}
\multicolumn{17}{@{}l}{\textit{Language-driven}} \\
Kling 2.5~\citep{kling} & \underline{0.31} & 6.93 & 43.37 & 0.315 & \underline{13.50} & 0.535 & \underline{0.380} & 38.13 & \textbf{99.78} & \textbf{87.87} & \textbf{97.57} & \underline{56.28} & 75.70 & 40.75 & \textbf{83.44} & 62.10 \\
Veo 3.1~\citep{veo} & 0.89 & 8.20 & 21.27 & 0.447 & 12.30 & 0.595 & 0.345 & 43.03 & 98.73 & 73.61 & \underline{95.42} & 55.49 & 76.57 & 32.15 & 79.96 & 56.06 \\
Hailuo 2.3~\citep{hailuo} & \textbf{0.28} & \textbf{5.18} & \textbf{54.61} & \textbf{0.510} & \textbf{13.56} & \textbf{0.518} & \textbf{0.388} & \textbf{49.00} & 98.94 & \underline{79.88} & 95.32 & \textbf{56.30} & 75.52 & \textbf{51.81} & \underline{81.19} & \textbf{66.50} \\
Wan 2.6 I2V~\citep{wan} & 0.38 & 6.12 & \underline{46.50} & \underline{0.485} & 13.29 & \underline{0.527} & 0.355 & \underline{46.88} & 99.27 & 69.43 & 94.13 & 53.93 & \textbf{77.53} & \underline{46.69} & 78.86 & \underline{62.78} \\
Seedance 1.5~\citep{seedance} & 0.78 & 7.78 & 28.47 & 0.370 & 12.29 & 0.601 & 0.343 & 39.01 & 95.96 & 56.60 & 94.82 & 54.45 & 74.89 & 33.74 & 75.34 & 54.54 \\
Vidu Q3~\citep{vidu} & 0.49 & 8.48 & 30.69 & 0.360 & 12.67 & 0.582 & 0.352 & 39.14 & 99.13 & 71.27 & 95.01 & 55.38 & 76.73 & 34.92 & 79.50 & 57.21 \\
HappyHorse 1.0~\citep{happyhorse} & 0.40 & \underline{5.55} & 45.66 & 0.415 & 12.45 & 0.564 & 0.351 & 42.16 & \underline{99.52} & 74.17 & 95.02 & 55.57 & \underline{76.99} & 43.91 & 80.25 & 62.08 \\
\bottomrule
\end{tabular*}%
}
\end{table*}

%% file: tables/interactive_results_da3.tex
\begin{table*}[t]
\centering
\small
\caption{\textbf{Dynamic-interaction track evaluation with DA3.}
We recompute Subject Control, Terrain Interaction, and affected aggregates with DA3; tasks evaluated using checklists remain unchanged, and aggregation and highlights follow \cref{tab:interactive-results}.}
\label{tab:interactive-results-da3}
\setlength{\heavyrulewidth}{0.096em}
\setlength{\lightrulewidth}{0.060em}
\setlength{\cmidrulewidth}{0.036em}
\resizebox{\linewidth}{!}{%
\setlength{\tabcolsep}{3pt}
\renewcommand{\arraystretch}{1.03}
\begin{tabular*}{1.20\linewidth}{@{\extracolsep{\fill}}l*{13}{c}@{}}
\toprule
\multirow[c]{2}{*}[-2pt]{\textbf{Model}} & \multirow[c]{2}{*}[-2pt]{\shortstack{\textbf{Subject}\\\textbf{Control}}} & \multirow[c]{2}{*}[-2pt]{\shortstack{\textbf{Terrain}\\\textbf{Inter.}}} & \multirow[c]{2}{*}[-2pt]{\shortstack{\textbf{Object}\\\textbf{Inter.}}} & \multirow[c]{2}{*}[-2pt]{\shortstack{\textbf{Social}\\\textbf{Inter.}}} & \multirow[c]{2}{*}[-2pt]{\shortstack{\textbf{Physical}\\\textbf{Reaction}}} & \multirow[c]{2}{*}[-2pt]{\shortstack{\textbf{Goal}\\\textbf{Completion}}} & \multirow[c]{2}{*}[-2pt]{\shortstack{\textbf{Subject}\\\textbf{Cons.}}} & \multirow[c]{2}{*}[-2pt]{\shortstack{\textbf{Motion}\\\textbf{Smooth.}}} & \multirow[c]{2}{*}[-2pt]{\shortstack{\textbf{Aesth.}\\\textbf{Quality}}} & \multirow[c]{2}{*}[-2pt]{\shortstack{\textbf{Imag.}\\\textbf{Quality}}} & \multicolumn{3}{c}{\textbf{Average}} \\
\cmidrule(lr){12-14}
& & & & & & & & & & & \textbf{Task} & \textbf{General} & \textbf{Overall} \\
\midrule
\rowcolor{gray!15}
\multicolumn{14}{@{}l}{\textit{Action-driven}} \\
WorldPlay~\citep{worldplay} & \underline{47.51} & \textbf{23.05} & \textbf{33.75} & \underline{51.40} & \underline{26.91} & -- & \underline{88.06} & \underline{98.09} & \underline{54.19} & \underline{68.76} & \underline{36.52} & \underline{77.28} & \underline{56.90} \\
LingBot-World~\citep{lingbot} & \textbf{53.43} & \underline{21.63} & \underline{25.94} & \textbf{60.37} & \textbf{33.43} & -- & \textbf{94.98} & \textbf{98.89} & \textbf{60.87} & \textbf{71.70} & \textbf{38.96} & \textbf{81.61} & \textbf{60.29} \\
\midrule
\rowcolor{gray!15}
\multicolumn{14}{@{}l}{\textit{Language-driven}} \\
Kling 2.5~\citep{kling} & 26.78 & 37.50 & 27.70 & 66.80 & 31.99 & 48.25 & \textbf{96.00} & \textbf{99.49} & 56.86 & 71.83 & 39.84 & \textbf{81.05} & 60.45 \\
Veo 3.1~\citep{veo} & \underline{35.12} & 42.19 & \textbf{75.96} & \textbf{85.10} & \underline{61.76} & \underline{85.30} & 92.07 & 99.12 & \textbf{58.21} & 72.65 & \textbf{64.24} & 80.51 & \textbf{72.38} \\
Hailuo 2.3~\citep{hailuo} & \textbf{35.16} & \underline{59.34} & 67.01 & 72.45 & \textbf{63.84} & 78.86 & 93.26 & \underline{99.31} & \underline{57.74} & 72.47 & 62.78 & 80.70 & 71.74 \\
Wan 2.6 I2V~\citep{wan} & 26.95 & 48.55 & 44.56 & 66.40 & 48.15 & 76.39 & \underline{94.46} & 98.15 & 56.30 & \textbf{74.71} & 51.83 & \underline{80.91} & 66.37 \\
Seedance 1.5~\citep{seedance} & 29.41 & 48.28 & 37.91 & 72.09 & 47.60 & 76.24 & 92.14 & 98.87 & 56.71 & 70.84 & 51.92 & 79.64 & 65.78 \\
Vidu Q3~\citep{vidu} & 26.15 & \textbf{62.56} & \underline{71.59} & \underline{81.91} & 61.23 & 78.26 & 92.78 & 98.76 & 57.24 & \underline{73.25} & \underline{63.62} & 80.51 & \underline{72.07} \\
HappyHorse 1.0~\citep{happyhorse} & 31.62 & 53.81 & 65.70 & 76.17 & 47.01 & \textbf{85.33} & 92.69 & 98.86 & 57.37 & 73.23 & 59.94 & 80.54 & 70.24 \\
\bottomrule
\end{tabular*}%
}
\end{table*}

%% file: sections/6_conclusion.tex
\section{Conclusion}
\label{sec:conclusion}

We presented \name, a unified hierarchical benchmark for diagnosing video world models beyond visual quality and explicit instruction fulfillment.
It distinguishes direct fulfillment of explicitly specified controls or targets from scene-conditioned reactions and detailed goal-directed execution that must be inferred from the initial scene.
This distinction is instantiated through four diagnostic levels, eight tasks, and 1{,}474 test cases.
Interface adaptation presents shared cases in the native formats of camera-, action-, and language-driven models, while the \emph{static-scene} and \emph{dynamic-interaction} tracks restrict evaluation to compatible interfaces rather than treating unsupported tasks as failures.

Evaluation of 20 representative models reveals a clear capability split.
Camera-driven models provide the most precise camera control and scene revisiting; action-driven models control subjects more precisely but often leave terrain, objects, nearby agents, and physical processes unresponsive; and language-driven models perform better on interaction and goal-directed tasks but follow composed controls less faithfully.
No evaluated model combines broad task coverage with consistently strong performance, showing that high visual quality and explicit instruction fulfillment do not guarantee \emph{inherent reactivity}.
Strong agreement between human and VLM checklist scores, together with stable model rankings under an alternative reconstruction backend, supports the reliability of these findings.

The current scope is bounded by the capabilities of available model interfaces.
Dynamic-interaction evaluation requires reliable third-person subject control, and Goal Completion remains limited to language-driven models.
Moreover, the metrics assess observable end-to-end behavior rather than determining where reasoning occurs or establishing that the video generator itself has learned an internal causal representation.
For closed-source commercial systems, proprietary prompt enhancement may contribute to scene grounding and execution planning.
Future extensions to broader interfaces, longer-horizon interactions, and more intervention-based settings would provide stronger tests of persistent world understanding.
Within its current scope, \name identifies where controllable video models succeed, where their generated worlds remain unresponsive, and which capabilities must be developed jointly.

%% file: sections/x_appendix.tex
\vspace{-4mm}
\section{Gallery}
\label{sec:appendix:gallery}

\begin{figure}[H]
  \centering
  \vspace{-6mm}
  \includegraphics[width=\linewidth,height=0.9\textheight,keepaspectratio]{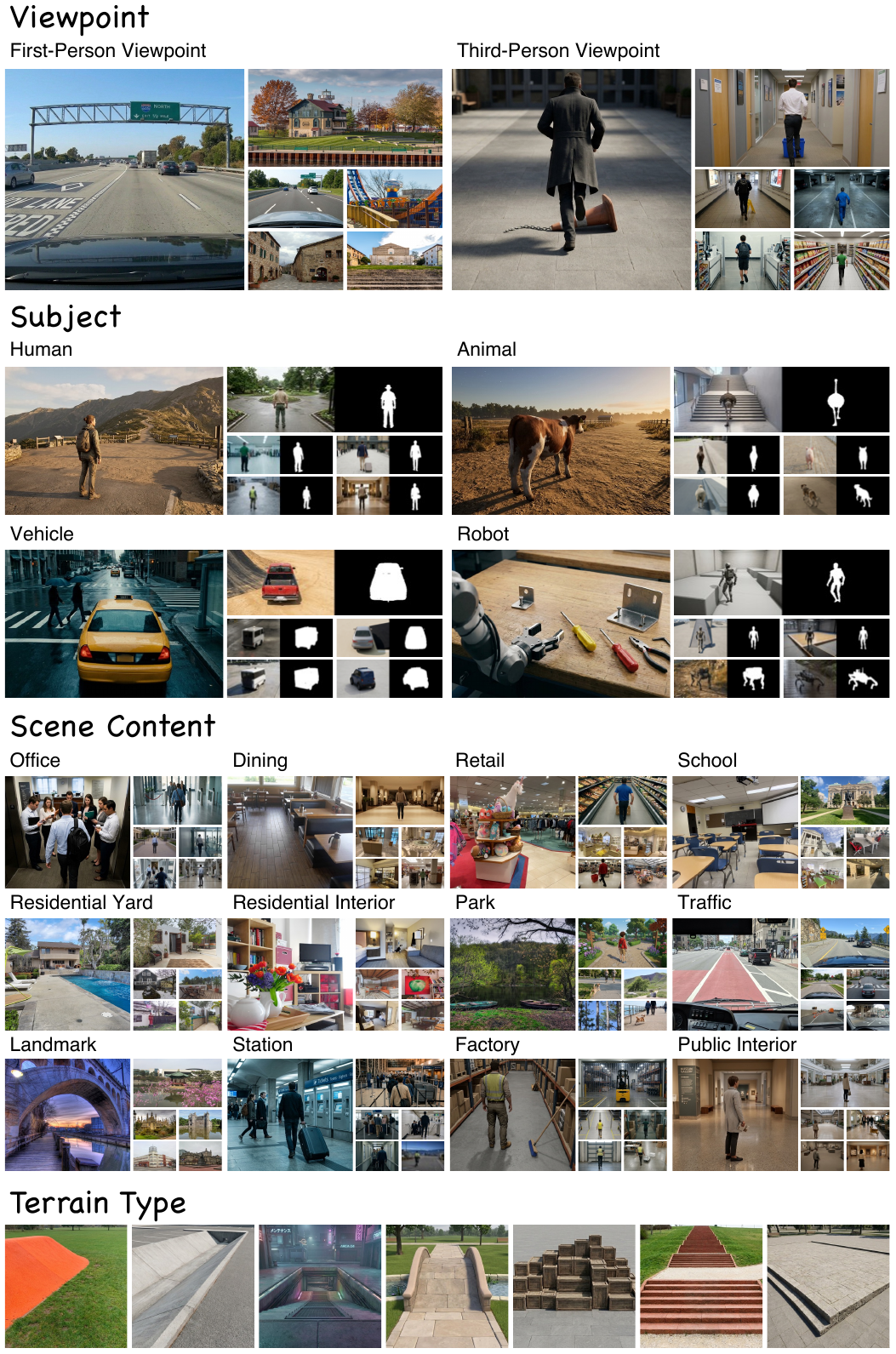}
  \vspace{-2mm}
  \caption{\textbf{\name Gallery.}
  Representative test cases span first- and third-person viewpoints; human, animal, vehicle, and robot subjects; diverse indoor and outdoor scene content; and a range of terrain types.}
  \label{fig:appendix-gallery}
\end{figure}

\section{Per-Model Inference Settings}
\label{sec:appendix:model-settings}

To preserve each model's native performance, we use its default resolution, video length, and other inference settings whenever applicable.
For streaming models with flexible generation lengths, we constrain the output to 100--200 frames to prevent quality drift in excessively long generations from biasing the evaluation.
\Cref{tab:model-generation-settings} reports the resulting resolution and frame count for each model.

Eligibility for the dynamic-interaction track requires reliable control of a visible third-person subject.
Among action-driven models, only WorldPlay~\citep{worldplay} and LingBot-World~\citep{lingbot} satisfy this requirement; the other five either do not support third-person subject control or cannot provide it reliably.
All evaluated language-driven models accept third-person subject-motion prompts, whereas camera-driven interfaces control only the camera.
Goal Completion is language-only within the dynamic-interaction track.

\input{tables/model_generation_settings}

\section{Qualitative Examples of the Eight Tasks}
\label{sec:appendix:task-examples}

\definecolor{taskteal}{HTML}{3C9C97}
\definecolor{tasknavy}{HTML}{214869}
\definecolor{taskmuted}{HTML}{66727E}
\definecolor{tasklight}{HTML}{F3F7F7}

\newcommand{\taskcasefigure}[4]{%
  \begin{minipage}{\linewidth}
    \centering
    \includegraphics[width=\linewidth]{#1}
    \captionof{figure}{\textbf{#2.} #3}
    \label{#4}
  \end{minipage}
}

\newcommand{\taskcasefigurecompact}[4]{%
  \begin{minipage}{\linewidth}
    \centering
    \includegraphics[width=0.86\linewidth]{#1}
    \captionsetup{skip=3pt}
    \captionof{figure}{\textbf{#2.} #3}
    \label{#4}
  \end{minipage}
}

\newtcolorbox{taskchecklist}{%
  enhanced,
  breakable,
  colback=tasklight,
  colframe=taskteal!55!white,
  borderline west={2.5pt}{0pt}{taskteal},
  boxrule=0.45pt,
  arc=1.5mm,
  left=8pt,
  right=8pt,
  top=6pt,
  bottom=6pt,
  before skip=7pt,
  after skip=2pt,
  title={Case-Specif{}ic Evaluation Checklist},
  coltitle=tasknavy,
  fonttitle=\sffamily\bfseries,
  fontupper=\small,
}

\newtcolorbox{taskchecklistcompact}[1][Case-Specif{}ic Evaluation Checklist]{%
  enhanced,
  breakable,
  colback=tasklight,
  colframe=taskteal!55!white,
  borderline west={2.5pt}{0pt}{taskteal},
  boxrule=0.45pt,
  arc=1.5mm,
  left=8pt,
  right=8pt,
  top=4pt,
  bottom=4pt,
  before skip=4pt,
  after skip=1pt,
  title={#1},
  coltitle=tasknavy,
  fonttitle=\sffamily\bfseries,
  fontupper=\small,
}

\newenvironment{taskcheckitems}{%
  \begin{itemize}[
    leftmargin=1.65em,
    label=\textcolor{taskteal}{\ding{113}},
    itemsep=0.30em,
    topsep=0.45em,
    parsep=0pt,
    partopsep=0pt
  ]
}{%
  \end{itemize}
}

\newenvironment{taskcheckitemscompact}{%
  \begin{itemize}[
    leftmargin=1.65em,
    label=\textcolor{taskteal}{\ding{113}},
    itemsep=0.08em,
    topsep=0.22em,
    parsep=0pt,
    partopsep=0pt
  ]
}{%
  \end{itemize}
}

The examples below instantiate all eight tasks in a shared visual format.
The highlighted image is the initial frame, followed by four temporally ordered generated frames.
The bottom panel shows the shared control intent or high-level goal.
For tasks evaluated using checklists, the checklist is shown only to explain the evaluation protocol and is withheld from the input.

\subsection{Camera Control}
\label{sec:appendix:camera-control}

Camera Control measures whether an ordered composition of camera motions follows the prescribed directions and temporal order without unintended drift.
In this case, the camera tilts down, pans left, and moves left.

\taskcasefigure
  {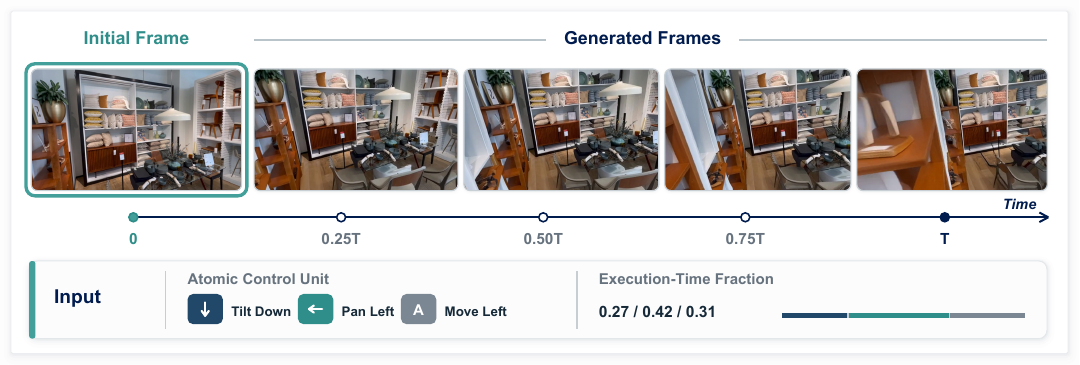}
  {Camera Control}
  {The three atomic camera controls occupy $0.27$, $0.42$, and $0.31$ of the video, respectively.}
  {fig:appendix-camera-control}

\subsection{Subject Control}
\label{sec:appendix:subject-control}

Subject Control applies atomic controls to a designated third-person subject.
Here, the subject should move forward, right, and left in order while remaining visually identifiable.

\taskcasefigure
  {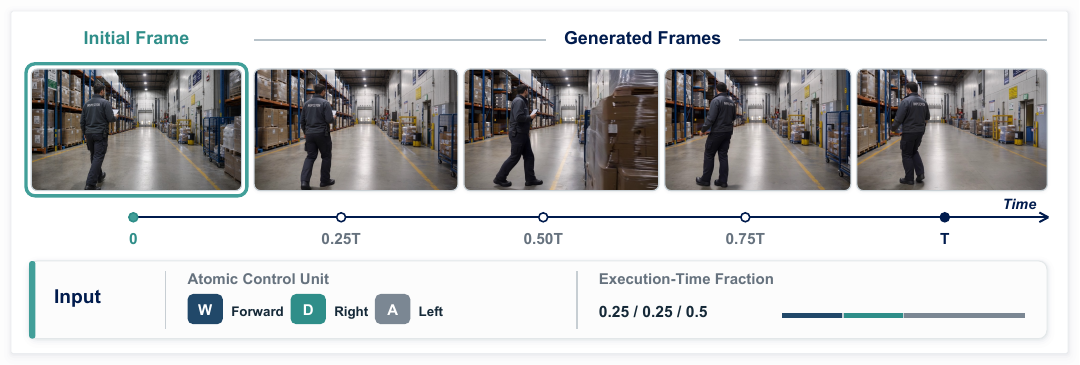}
  {Subject Control}
  {The Forward, Right, and Left controls occupy $0.25$, $0.25$, and $0.50$ of the video.}
  {fig:appendix-subject-control}

\subsection{Scene Revisit}
\label{sec:appendix:scene-revisit}

Scene Revisit couples a round-trip camera motion with a spatial-memory requirement: the camera should return to the initial viewpoint while preserving the scene.

\taskcasefigure
  {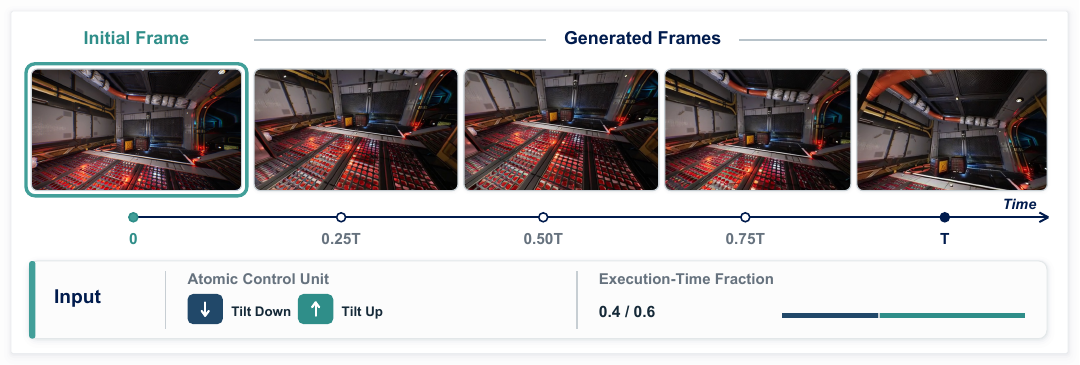}
  {Scene Revisit}
  {The camera first tilts down and then tilts up; success requires both execution of the motion and recovery of a consistent revisited view.}
  {fig:appendix-scene-revisit}

\subsection{Terrain Interaction}
\label{sec:appendix:terrain-interaction}

Terrain Interaction specifies only horizontal subject motion.
The model must infer the vertical adaptation required by the visible terrain---in this case, traversing the stairs while continuing forward.

\taskcasefigure
  {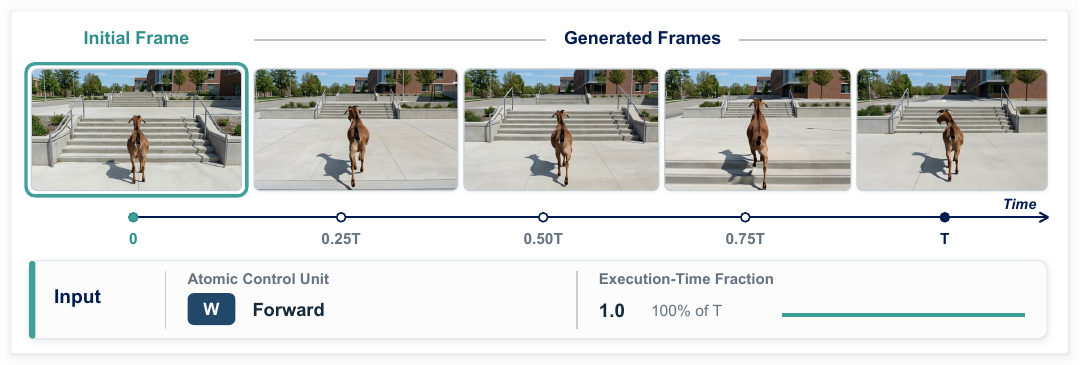}
  {Terrain Interaction}
  {The Forward control is active for the full video, while the stair-climbing response is implied by the scene rather than stated in the input.}
  {fig:appendix-terrain-interaction}

\subsection{Object Interaction}
\label{sec:appendix:object-interaction}

Object Interaction tests whether contact with a designated target causes a response consistent with the target's physical type.
Here, the worker pushes a bus cart forward into a lightweight sign.

\taskcasefigure
  {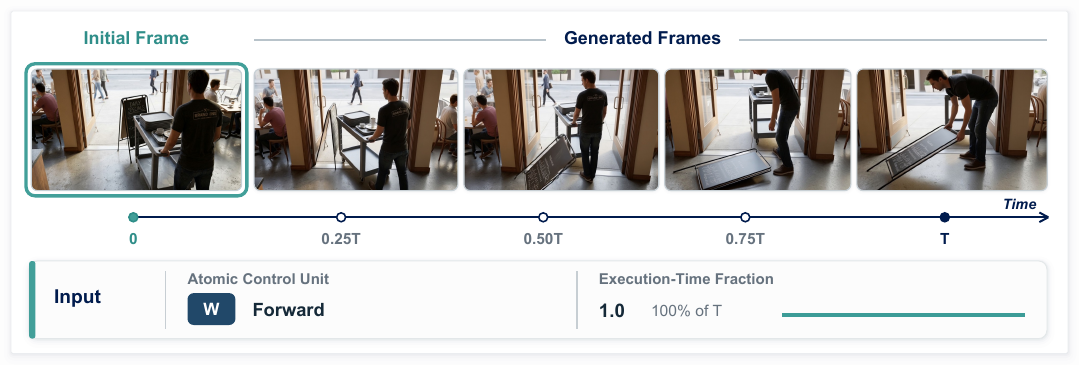}
  {Object Interaction}
  {Only the Forward control is specified; the sign's contact response is withheld from the model input and assessed with the checklist below.}
  {fig:appendix-object-interaction}

\begin{taskchecklist}
\textcolor{taskmuted}{\emph{Evaluation only; not part of the model-facing input.}}
\begin{taskcheckitems}
  \item Does the worker keep the bus cart rolling straight forward on the same approach line into the sign without steering away before impact?
  \item Does the bus cart make the first contact with the sign, specifically low on the sign's lower frame, panel edge, or legs, rather than the worker's body hitting it directly?
  \item After contact, does the A-frame sign behave like a light freestanding hinged sign by tipping, folding, sliding, or getting shoved aside?
  \item Does the sign remain visibly unattached and mobile, rather than acting as if fixed rigidly to the floor or doorway?
  \item Does the sign's motion follow the cart's low forward push, with the base and legs reacting first instead of the top moving in an implausible independent way?
  \item Does the worker stay behind the cart with a continued forward pushing posture through the moment of impact?
  \item If the cart continues through the doorway area, is that continuation enabled by the sign being displaced or collapsing aside rather than unrealistically blocking the cart like a solid barrier?
  \item Does the continuation keep the cart--sign contact zone and the sign's passive response visible enough to judge the impact clearly?
\end{taskcheckitems}
\end{taskchecklist}

\subsection{Social Interaction}
\label{sec:appendix:social-interaction}

Social Interaction evaluates whether nearby agents respond plausibly when a controlled subject enters their social or safety space.
Here, a sedan enters a storefront crossing with two pedestrians and a shopping cart.

\taskcasefigure
  {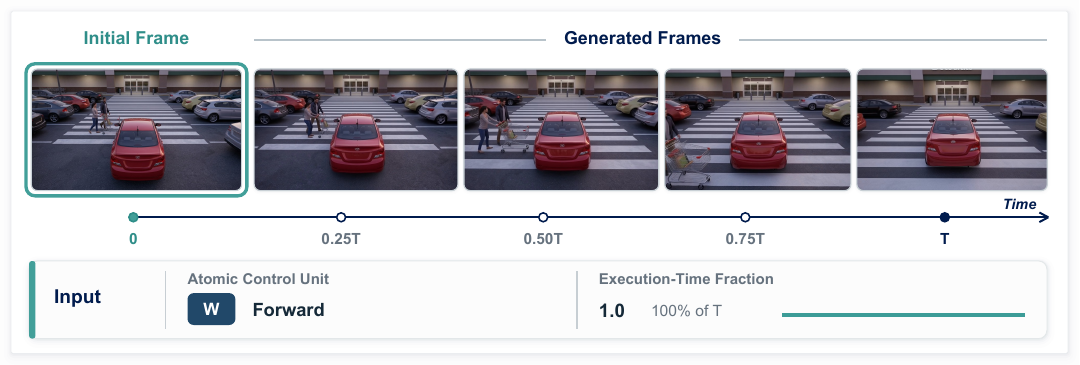}
  {Social Interaction}
  {Only the car's Forward control is specified; the checklist evaluates the pedestrians' response.}
  {fig:appendix-social-interaction}

\begin{taskchecklistcompact}[Social Interaction Evaluation Checklist]
\textcolor{taskmuted}{\emph{Evaluation only; not part of the model-facing input.}}
\begin{taskcheckitemscompact}
  \item Does the red sedan continue inching forward into the storefront crossing?
  \item Do the cart pusher and companion remain visible enough to judge how they pass the car nose?
  \item Does the cart path or walking pace visibly change near the sedan's protruding front end?
  \item Is the cart interaction focused on the crossing area directly in front of the car?
  \item Does the continuation stay at ordinary parking-lot speed and behavior?
  \item Does the sedan remain the dominant source of social pressure in the scene?
  \item Do the pedestrians guiding the shopping cart respond plausibly as they move around the vehicle nose?
  \item Does the clip avoid unrelated dominant events such as another car cutting through the scene?
\end{taskcheckitemscompact}
\end{taskchecklistcompact}

\subsection{Physical Reaction}
\label{sec:appendix:physical-reaction}

Physical Reaction evaluates the temporal development of a physical process, not merely whether contact occurs.
Here, the woman remains stationary while a towel-loaded laundry basket, whose center of mass overhangs the washer edge, begins to tip and fall.

\taskcasefigure
  {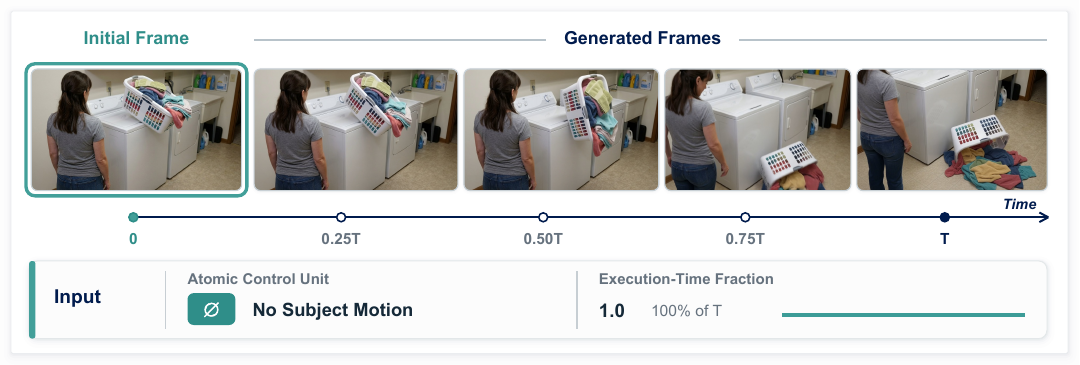}
  {Physical Reaction}
  {The $\varnothing$ (stop) control specifies no subject motion; the basket's rotation about the support edge and subsequent fall should unfold autonomously.}
  {fig:appendix-physical-reaction}

\begin{taskchecklistcompact}[Physical Reaction Evaluation Checklist]
\textcolor{taskmuted}{\emph{Evaluation only; not part of the model-facing input.}}
\begin{taskcheckitemscompact}
  \item Does the woman remain stationary throughout the short continuation?
  \item Does the basket continue rotating outward and downward from its already tilted position?
  \item Does the basket fall off the washing machine rather than sliding back to a fully supported position on top?
  \item Does the motion begin as a tip about the washer edge or remaining contact point, consistent with the overhanging weight pulling it over?
  \item Does the overhanging towel load move with the basket and contribute to the same outward fall direction?
  \item Does the basket fall without any new touch, grab, or bump from the woman?
  \item After leaving support, do the basket and laundry move downward into the open space in front of the machine rather than floating or reversing upward?
\end{taskcheckitemscompact}
\end{taskchecklistcompact}

\subsection{Goal Completion}
\label{sec:appendix:goal-completion}

Goal Completion provides a high-level goal instead of an atomic control sequence.
The model must ground the goal in the initial scene and generate coherent execution steps toward the desired target.

\taskcasefigure
  {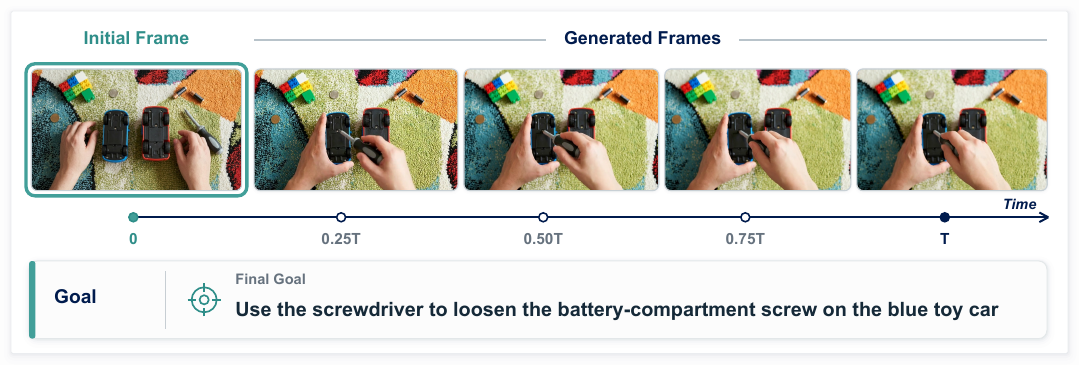}
  {Goal Completion}
  {The goal is to loosen the blue toy car's battery-compartment screw with the screwdriver.}
  {fig:appendix-goal-completion}

\begin{taskchecklistcompact}[Goal Completion Evaluation Checklist]
\textcolor{taskmuted}{\emph{Evaluation only; not part of the model-facing input.}}
\begin{taskcheckitemscompact}
  \item Does the person pick up and use the visible screwdriver?
  \item Does the action target the blue toy car rather than the red toy vehicle?
  \item Does the screwdriver tip align with the blue car's battery-compartment screw?
  \item Does the continuation show a clear screw-loosening motion on the blue car?
  \item Does the red toy vehicle remain unused throughout the task?
  \item Does the person avoid using the nearby coin as a tool?
  \item Does the blue car's battery cover end visibly loosened or ready to open?
\end{taskcheckitemscompact}
\end{taskchecklistcompact}

%% file: tables/model_generation_settings.tex
\begingroup
\newcommand{\dynyes}{\textcolor{green!55!black}{\ding{51}}}
\newcommand{\dynno}{\textcolor{red!70!black}{\ding{55}}}
\begin{table}[h]
\centering
\small
\caption{\textbf{Per-model inference settings and eligibility for the dynamic-interaction track.}
Resolution is the center-cropped input size in width$\times$height order, and Frames reports the evaluated video length.
\emph{TPV} and \emph{FPV} denote third- and first-person viewpoints, respectively.
\dynyes{} denotes eligibility for the dynamic-interaction track;
\dynno{} denotes FPV-only control or unreliable TPV subject control.
\textsuperscript{\ensuremath{\dagger}} denotes unreliable TPV subject control.
A dash denotes that the track is not applicable because the model controls only the camera.}
\label{tab:model-generation-settings}
\setlength{\tabcolsep}{11pt}
\renewcommand{\arraystretch}{1.08}
\begin{tabular}{@{}lcccc@{}}
\toprule
\textbf{Model} & \textbf{Backend} & \textbf{Resolution} & \textbf{Frames} &
\textbf{Dynamic Track} \\
\midrule
\rowcolor{gray!15}
\multicolumn{5}{@{}l}{\textit{Camera-driven}} \\
TrajectoryCrafter~\citep{trajectorycrafter} & Local & $672{\times}384$ & 49 & -- \\
ReCamMaster~\citep{recammaster} & Local & $832{\times}480$ & 81 & -- \\
Voyager~\citep{voyager} & Local & $768{\times}512$ & 49 & -- \\
FantasyWorld~\citep{fantasyworld} & Local & $592{\times}336$ & 81 & -- \\
NeoVerse~\citep{neoverse} & Local & $560{\times}336$ & 81 & -- \\
InSpatio-World (1.3B)~\citep{inspatio} & Local & $832{\times}480$ & 81 & -- \\
\midrule
\rowcolor{gray!15}
\multicolumn{5}{@{}l}{\textit{Action-driven}} \\
Hunyuan-GameCraft~\citep{gamecraft} & Local & $1216{\times}704$ & 132 & \dynno{} TPV\textsuperscript{\ensuremath{\dagger}} \\
Astra~\citep{astra} & Local & $832{\times}480$ & 161 & \dynno{} FPV only \\
WorldPlay~\citep{worldplay} & Local & $832{\times}480$ & 125 & \dynyes{} TPV \\
Yume 1.5~\citep{yume15} & Local & $1280{\times}704$ & 145 & \dynno{} FPV only \\
LingBot-World~\citep{lingbot} & Local & $832{\times}464$ & 161 & \dynyes{} TPV \\
Infinite-World~\citep{infiniteworld} & Local & $896{\times}448$ & 161 & \dynno{} FPV only \\
Matrix-Game 3.0~\citep{matrixgame3} & Local & $1280{\times}704$ & 177 & \dynno{} FPV only \\
\midrule
\rowcolor{gray!15}
\multicolumn{5}{@{}l}{\textit{Language-driven}} \\
Kling 2.5~\citep{kling} & API & $1280{\times}720$ & 153 & \dynyes{} TPV prompt \\
Veo 3.1~\citep{veo} & API & $1280{\times}720$ & 192 & \dynyes{} TPV prompt \\
Hailuo 2.3~\citep{hailuo} & API & $1024{\times}768$ & 141 & \dynyes{} TPV prompt \\
Wan 2.6 I2V~\citep{wan} & API & $1280{\times}720$ & 150 & \dynyes{} TPV prompt \\
Seedance 1.5~\citep{seedance} & API & $1280{\times}720$ & 97 & \dynyes{} TPV prompt \\
Vidu Q3~\citep{vidu} & API & $1280{\times}720$ & 121 & \dynyes{} TPV prompt \\
HappyHorse 1.0~\citep{happyhorse} & API & $1280{\times}720$ & 123 & \dynyes{} TPV prompt \\
\bottomrule
\end{tabular}
\end{table}
\endgroup